\def\year{2021}\relax
\documentclass[letterpaper]{article} 
\usepackage{aaai21}  
\usepackage{times}  
\usepackage{helvet} 
\usepackage{courier}  
\usepackage[hyphens]{url}  
\usepackage{graphicx} 
\usepackage{xcolor}
\usepackage{natbib}  
\usepackage{caption} 
\usepackage[utf8]{inputenc} 
\usepackage[T1]{fontenc}    
\usepackage{url}            
\usepackage{booktabs}       
\usepackage{amsfonts}       
\usepackage{nicefrac}       
\usepackage{microtype}      
\usepackage{graphicx}
\usepackage{graphics}
\usepackage{nccmath}
\usepackage{stackengine}
\usepackage{algorithm}
\usepackage{algpseudocode}
\usepackage{enumitem}
\usepackage{bbm}
\usepackage{placeins}
\usepackage{amsmath,amssymb,amsfonts,amsthm,nccmath}
\DeclareMathOperator*{\argmin}{argmin}
\DeclareMathOperator*{\argmax}{argmax}
\newtheorem{definition}{Definition}

\newtheorem{problem}{Problem}
\newtheorem{proposition}{Proposition}

\usepackage[toc,page, title, titletoc]{appendix}
\usepackage[symbol]{footmisc}

\title{Minimax Regret Optimisation for Robust Planning in Uncertain Markov Decision Processes}

\author{
	Marc Rigter,
	Bruno Lacerda,
	Nick Hawes
	\\
}
\affiliations{
	Oxford Robotics Institute, University of Oxford, United Kingdom\\
	
	\{mrigter, bruno, nickh\}@robots.ox.ac.uk
	
}

\begin{document}

\maketitle


\begin{abstract}
The parameters for a Markov Decision Process (MDP) often cannot be specified exactly.
Uncertain MDPs (UMDPs) capture this model ambiguity by defining sets which the parameters belong to. 
Minimax regret has been proposed as an objective for planning in UMDPs to find robust policies which are not overly conservative. 
In this work, we focus on planning for Stochastic Shortest Path (SSP) UMDPs with uncertain cost and transition functions.
We introduce a Bellman equation to compute the regret for a policy. 
We propose a dynamic programming algorithm that utilises the regret Bellman equation, and show that it optimises minimax regret exactly for UMDPs with independent uncertainties.
For coupled uncertainties, we extend our approach to use options to enable a trade off between computation and solution quality.
We evaluate our approach on both synthetic and real-world domains, showing that it significantly outperforms existing baselines.
\end{abstract}


\section{Introduction}
Markov Decision Processes (MDPs) are a powerful tool for sequential decision making in stochastic domains. 
However, the parameters of an MDP are often estimated from limited data, and therefore cannot be specified exactly~\cite{lacerda2019probabilistic, moldovan2012risk}. 
By disregarding model uncertainty and planning on the estimated MDP, performance can be much worse than anticipated~\cite{mannor2004bias}.

For example, consider an MDP model for medical decision making, where transitions correspond to the stochastic health outcomes for a patient as a result of different treatment options.
An estimated MDP model for this problem can be generated from observed data~\cite{schaefer2005modeling}.
However, such a model does not capture the variation in transition probabilities due to patient heterogeneity: any particular patient may respond differently to treatments than the average due to unknown underlying factors.
Additionally, such a model does not consider uncertainty in the model parameters due to limited data.
As a result, Uncertain MDPs (UMDPs) have been proposed as more suitable model for domains such as medical decision making~\cite{zhang2017robust} where the model cannot be specified exactly.

UMDPs capture model ambiguity by defining an uncertainty set in which the true MDP cost and transition functions lie.
In this work, we address offline planning for UMDPs.
Most research in this setting has focused on optimising the expected value for the worst-case MDP parameters using robust dynamic programming~\cite{iyengar2005robust,nilim2005robust}. 
However, this can result in overly conservative policies which perform poorly in the majority of possible scenarios~\cite{delage2010percentile}.
\textit{Minimax regret} has been proposed as alternative metric for robust planning which is less conservative~\cite{regan2009regret,xu2009parametric}.
The aim is to find the policy with the minimum gap between its expected value and the optimal value over all possible instantiations of model uncertainty. 
However, optimising minimax regret is challenging and existing methods do not scale well.

In this work, we introduce a Bellman equation to decompose the computation of the regret for a policy into a dynamic programming recursion.
We show that if uncertainties are \textit{independent}, we can perform minimax value iteration using the regret Bellman equation to efficiently optimise minimax regret exactly.\footnote[1]{Following publication, the authors realised that this claim is incorrect. Please see the corrigendum.} 
To our knowledge, this is the first scalable exact algorithm for minimax regret planning in UMDPs with both uncertain cost and transition functions.
To address problems with \textit{dependent} uncertainties, we introduce the use of \emph{options}~\cite{sutton1999between} to capture dependence over sequences of $n$ steps. 
By varying $n$, the user may trade off computation time against solution quality.

Previous works have addressed regret-based planning in finite horizon problems~\cite{ahmed2013regret, ahmed2017sampling}, or problems where there is only uncertainty in the cost function~\cite{regan2009regret,regan2010robust, regan2011robust, xu2009parametric}. 
We focus on the more general problem of Stochastic Shortest Path (SSP) UMDPs with uncertain cost and transition functions.
The main contributions of this work are: 
\begin{itemize}
	\itemsep-1.5pt
	\item Introducing a Bellman equation to compute the regret for a policy using dynamic programming.
	\item An efficient algorithm to optimise minimax regret exactly in models with independent uncertainties by performing minimax value iteration using our novel Bellman equation.$^*$
	\item Proposing the use of options to capture dependencies between uncertainties to trade off solution quality against computation for models with dependent uncertainties.
\end{itemize}
Experiments in both synthetic and real-world domains demonstrate that our approach considerably outperforms existing baselines.


\section{Related Work}
The worst-case expected value for a UMDP can be optimised efficiently with robust dynamic programming provided that the uncertainty set is convex, and the uncertainties are independent between states~\cite{iyengar2005robust, nilim2005robust, wiesemann2013robust}.
However, optimising for the worst-case expected value often results in overly conservative policies~\cite{delage2010percentile}.
This problem is exacerbated by the independence assumption which allows all parameters to be realised as their worst-case values simultaneously. 
Sample-based UMDPs represent model uncertainty with a finite set of possible MDPs, capturing dependencies between uncertainties~\cite{adulyasak2015solving, ahmed2013regret, ahmed2017sampling, chen2012tractable, cubuktepe2020scenario, steimle2018multi}. 
For sample-based UMDPs, dependent uncertainties can also be represented by augmenting the state space~\cite{mannor2016robust}, however this greatly enlarges the state space even for a modest number of samples.

To compute less conservative policies, alternative planning objectives to worst-case expected value have been proposed. 
Possibilities include forgoing robustness and optimising average performance~\cite{adulyasak2015solving,steimle2018multi},
performing chance-constrained optimisation under a known distribution of model parameters~\cite{delage2010percentile},
and computing a Pareto-front for multiple objectives~\cite{scheftelowitsch2017multi}. 
\textit{Minimax regret} has been proposed as an intuitive objective which is less conservative than optimising for the worst-case expected value~\cite{xu2009parametric}, but can be considered robust as it optimises worst-case sub-optimality.
Minimax regret in UMDPs where only the cost function is uncertain is addressed in~\cite{regan2009regret, regan2010robust, regan2011robust, xu2009parametric}. 

Limited research has addressed minimax regret in UMDP planning with \emph{both} uncertain cost and transition functions.
For sample-based UMDPs, the best stationary policy can be found by solving a Mixed Integer Linear Program (MILP), however this approach does not scale well~\cite{ahmed2013regret}.
A policy-iteration algorithm is proposed by~\citet{ahmed2017sampling} to find a policy with locally optimal minimax regret. 
However, this approach is only suitable for finite-horizon planning in which states are indexed by time step and the graph is acyclic.
An approximation proposed by~\citet{ahmed2013regret} optimises minimax Cumulative Expected Myopic Regret (CEMR). CEMR myopically approximates regret by comparing local actions, rather than evaluating overall performance.
Our experiments show that policies optimising CEMR often perform poorly for minimax regret.
Unlike CEMR, our approach optimises minimax regret exactly for problems with independent uncertainties.

Regret is used to measure performance in reinforcement learning (RL) (eg.~\citealt{jaksch2010near, cohen2020near, tarbouriech2020no}).
In the RL setting, the goal is to minimise the \textit{total} regret, which is the total loss incurred throughout training over many episodes.
In contrast, in our UMDP setting we plan offline to optimise the worst-case regret for a policy.
This is the regret for a fixed policy evaluated over a single episode, assuming the MDP parameters are chosen adversarially.
In RL, options~\cite{sutton1999between} have been utilised for learning robust policies with temporally extended actions~\cite{mankowitz2018learning}.
In this work, we use options to capture dependencies between model uncertainties throughout the execution of each option.

Another approach to address MDPs which are not known exactly is Bayesian RL~\cite{ghavamzadeh2015bayesian} which adapts the policy online throughout execution.
In contrast to our setting, Bayesian RL typically does not address worst-case performance and requires access to a distribution over MDPs rather than a set. 
The offline minimax regret setting we consider is more appropriate for safety-critical domains such as medical decision making, where the policy must be scrutinised by regulators prior to deployment, and robustness to worst-case suboptimality is important.
\vspace{-1mm}


\section{Preliminaries}
\begin{definition}
	\label{def:ssp}
	An SSP MDP is defined as a tuple $\mathcal{M}=(S, s_0, A, C, T, G)$.
	$S$ is the set of states, $s_0 \in S$ is the initial state, $A$ is the set of actions, $C: S \times A \times S \rightarrow {\rm I\!R}$ is the cost function, and $T: S \times A \times S \rightarrow [0, 1]$ is the transition function. 
	$G \subset S$ is the set of goal states. Each goal state $s_g \in G$ is absorbing and incurs zero cost. 
\end{definition}
The expected cost of applying action $a$ in state $s$ is $\bar{C}(s, a) =\sum_{s' \in S} T(s, a, s') \cdot C(s, a, s')$.
The minimum expected cost at state $s$ is $\bar{C}^*(s) = \min_{a \in A} \bar{C}(s,a)$.
A finite \textit{path} is a finite sequence of states visited in the MDP.
%
%
%
A \textit{history-dependent} policy maps finite paths to a distribution over action choices. 
A \textit{stationary} policy only considers the current state. 
%
%
A policy is \textit{deterministic} if it chooses a single action at each step.
The set of all policies is denoted $\Pi$.
A policy is \textit{proper at} $s$ if it reaches $s_g \in G$ from $s$ with probability 1. 
A policy is \textit{proper} if it is proper at all states.
In an SSP MDP, the following assumptions are made~\cite{kolobov2012planning}:
a) there exists a proper policy, and b) every improper policy incurs infinite cost at all states where  it is improper.

In this work, we aim to minimise the regret for a fixed policy over a single episode which is defined as follows.

\begin{definition}
\label{def:reg_pol}
The regret for a policy $\pi \in \Pi$, denoted $reg(s_0, \pi)$, is defined as
\vspace{-1mm}
\small
\begin{equation}
reg(s_0, \pi) = V(s_0, \pi) - V(s_0, \pi^*),
\end{equation}
\normalsize
\end{definition}
\vspace{-1mm}
\noindent where $V(s, \pi)$ is the value of a policy $\pi$ in state $s$ according to the following Bellman equation,
\vspace{-1mm}
\small
\begin{equation}
V(s, \pi) = \sum_{a \in A} \pi(s, a) \cdot  [\bar{C}(s,a) + \sum_{s' \in S}T(s, a, s') \cdot V(s', \pi)] ,
\label{eq:bellman}
\vspace{-1mm}
\end{equation}

\normalsize
\noindent and $\pi^*$ is the policy with minimal expected value. Intuitively, the regret for a policy is the expected suboptimality over a single episode. ~\citet {ahmed2013regret, ahmed2017sampling} proposed Cumulative Expected Myopic Regret (CEMR) as a regret approximation.

\begin{definition}
The CEMR of policy $\pi$ at state $s$, denoted $cemr(s, \pi)$ is defined as

\vspace{-2mm}
\small
\begin{multline}
\label{eqn:cer}
cemr(s, \pi) = {\textstyle\sum}_{a \in A} \pi(s, a) \cdot \\ [\bar{C}(s,a) -  \bar{C}^*(s) + {\textstyle\sum}_{s' \in S}T(s, a, s') \cdot cemr(s', \pi)]. 
\end{multline}
\normalsize
\end{definition}
\vspace{-1mm}

$\bar{C}(s,a) -  \bar{C}^*(s)$ is the gap between the expected cost of $a$, and the best expected cost for any action at $s$. CEMR is myopic, accumulating the local regret relative to the actions available at each state.

\vspace{-1mm}
\subsubsection{Uncertain MDPs}
\vspace{-0.5mm}
We use the commonly employed sample-based UMDP definition~\cite{adulyasak2015solving, ahmed2013regret, ahmed2017sampling, chen2012tractable, steimle2018multi}. This representation captures dependencies between uncertainties because each sample represents an entire MDP. As we are interested in worst-case regret, we  only require  samples which provide adequate coverage over possible MDPs, rather than a distribution over MDPs.

\begin{definition}
	An SSP UMDP is defined by the tuple $(S, s_0, A, \mathcal{C}, \mathcal{T}, G)$.
	$S$, $s_0$, $A$, and $G$ are defined as for SSP MDPs.
	${\mathcal{T} = \{T_1, T_2,\ldots, T_{|\xi|}\}}$ denotes a finite  set of possible transition functions and ${\mathcal{C} = \{C_1, C_2,\ldots, C_{|\xi|} \}}$ denotes the associated set of possible cost functions. 
	A sample of model uncertainty, $\xi_q$, is defined as ${\xi_q = (C_q, T_q)}$ where ${C_q \in \mathcal{C}}$, ${T_q \in \mathcal{T}}$.
	The set of samples is denoted $\xi$.
\end{definition}

\vspace{-0.5mm}
We provide a definition for independent uncertainty sets, equivalent to the state-action rectangularity property introduced in~\cite{iyengar2005robust}. Intuitively this means that uncertainties are decoupled between subsequent action choices.

%
%

\begin{definition}
\label{def:independence}
    Set $\mathcal{T}$  is  independent over state-action pairs if $\mathcal{T} = \times_{(s, a) \in S \times A} \mathcal{T}^{s, a} $
    where $\mathcal{T}^{s, a}$ is the set of possible distributions over $S$ after applying $a$ in $s$, and $\times$ denotes the Cartesian product.
\end{definition}

\vspace{-0.5mm}
The definition of independence for cost functions is analogous.
In this work we wish to find $\pi_{reg}$, the policy which minimises the maximum regret over the uncertainty set.

\begin{problem}
\label{prob:1}
    Find the minimax regret policy defined as 
    \vspace{-1mm}
    \begin{equation}
        \pi_{reg}  = \argmin_{\pi \in \Pi} \max_{\xi_q \in \xi} reg_q(s_0, \pi),
    \end{equation}
	\vspace{-4mm}
\end{problem}

\noindent where $reg_q(s_0, \pi)$ is the regret of $\pi$ in the MDP corresponding to sample $\xi_q$.
In general, stochastic policies are required to hedge against alternate possibilities~\cite{xu2009parametric}, and history-dependent policies are required if uncertainties are dependent~\cite{steimle2018multi, wiesemann2013robust}. 
If only stationary deterministic policies are considered, a minimax regret policy can be computed exactly by solving a MILP~\cite{ahmed2013regret}.
An approximation for minimax regret is to find the policy with minimax CEMR~\cite{ahmed2013regret, ahmed2017sampling}:

\vspace{-2mm}
\begin{equation}
     \pi_{cemr} = \argmin_{\pi \in \Pi} \max_{\xi_q \in \xi} cemr_q(s_0, \pi),
\end{equation}
\vspace{-2mm}
   
\noindent where $cemr_q(s_0, \pi)$ is the CEMR of $\pi$ corresponding to $\xi_q$.

Our work is closely connected to the UMDP solution which finds the best expected value for the worst-case parameters~\cite{iyengar2005robust, nilim2005robust, wiesemann2013robust}.
We refer to the resulting policy as the \textit{robust} policy.
Assuming independent uncertainties per Def.~\ref{def:independence}, finding the robust policy can be posed as a Stochastic Game (SG) between the agent, and an adversary $\sigma^1 : S \times A \times \xi \rightarrow \{0, 1\}$ which responds to the action of the agent by applying the worst-case parameters at each step:

\vspace{-2mm}
\begin{equation}
\label{eq:robust}
    \pi_{robust} = \argmin_{\pi \in \Pi} \max_{\sigma^1} V(s_0, \pi).
    \vspace{-2mm}
\end{equation}

\noindent The meaning of the superscript for $\sigma^1$ will become clear later.

For this problem, the optimal value function may be found via minimax Value Iteration (VI)
and corresponds to a deterministic stationary policy for both players~\cite{wiesemann2013robust}.
For SSPs, convergence is guaranteed if:
a) there exists a policy for the agent which is proper for all possible policies of the adversary, and b) for any states where $\pi$ and $\sigma$ are improper, the expected cost for the agent is infinite~\cite{patek1999stochastic}.


\vspace{-2mm}
\section{Regret Bellman Equation}
\vspace{-0.5mm}
Our first contribution is Proposition~\ref{prop:creg} which introduces a Bellman equation to compute the regret for a policy via dynamic programming. Full proofs of all propositions are in the full version of the paper~\cite{rigter2020minimax}.

\begin{proposition}
	\label{prop:creg}
	(Regret Bellman Equation) The regret for a proper policy, $\pi$, can be computed via the following recursion
	\vspace{-5mm}
	\small
	\begin{multline}
		\label{eq:creg}
		reg(s, \pi) = {\textstyle\sum}_{a \in A} \pi(s, a) \cdot \\ \big[Q^{gap}(s, a) + \textstyle{\sum}_{s' \in S}T(s, a, s')\cdot reg(s', \pi) \big], \hspace{5pt} \normalsize{\textit{where}}
	\end{multline}
	\vspace{-5mm}
	
	\begin{equation}
		\label{eq:qgap}
		Q^{gap}(s, a) = \big[ \bar{C}(s, a) + \textstyle{\sum}_{s' \in S}T(s, a, s') \cdot V(s', \pi^*) \big] - V(s, \pi^*),
	\end{equation}
	\normalsize
	and  $reg(s, \pi) = 0,\  \forall s \in G$. 
\end{proposition}
\noindent $Q^{gap}$ represents the suboptimality attributed to an $s,a$ pair.

\textit{Proof sketch}: unrolling Eq.~\ref{eq:creg}-\ref{eq:qgap} from $s_0$ for $h$ steps we have
\vspace{-3mm}
\scriptsize
\begin{multline*}
	reg(s_0, \pi) = - V(s_0, \pi^*) +
	\sum_{a} \pi(s_0, a) \bigg[\bar{C}(s_0, a)  + \\
	\sum_{s'}T(s_0, a, s') \bigg[\sum_{a}\pi(s', a) \bigg[\bar{C}(s', a)  + \ldots  
	\bigg[ \sum_{a} \pi(s^{h\textnormal{-}1}, a) \bigg[ \bar{C}(s^{h\textnormal{-}1}, a) \\ +
	\sum_{s^{h}}T(s^{h\textnormal{-}1}, a, s^{h}) V(s^h, \pi^*)
	+  \sum_{s^h}T(s^{h\textnormal{-}1}, a, s^h)reg(s^h, \pi) \bigg] \bigg] \ldots \bigg] \bigg] \bigg]
\end{multline*}
\normalsize
\vspace{-1mm}

Taking $h \rightarrow \infty$ we have $s^h \in G$ under the definition of a proper policy. Thus, $V(s^h, \pi^*) = reg(s^h, \pi) = 0$. Simplifying, we get $reg(s_0, \pi) = V(s_0, \pi) - V(s_0, \pi^*)$ which is the original definition for the regret of a policy. \qedsymbol

\vspace{-1mm}

\section{Minimax Regret Optimisation}
In this section, we describe how Proposition~\ref{prop:creg} can be used to optimise minimax regret in UMDPs. We separately address UMDPs with independent and dependent uncertainties.

\vspace{-2mm}
\subsection{Exact Solution for Independent Uncertainties}
\vspace{-0.5mm}
 To address minimax regret optimisation in UMDPs with independent uncertainties per Def.~\ref{def:independence}, we start by considering the following SG in Problem~\ref{prob:creg_n1}.
 At each step, the agent chooses an action, and the adversary $\sigma^1: S \times A \times \xi \rightarrow \{0, 1\} $ reacts to this choice by choosing the MDP sample to be applied for that step to maximise the regret of the policy.

 \begin{problem}
 \label{prob:creg_n1}
 Find the minimax regret policy in the stochastic game defined by
 \vspace{-1.5mm}
 \begin{equation}
\label{eq:creg_decoupled_n1}
    \pi_{reg}^1 = \argmin_{\pi \in \Pi} \max_{\sigma^1} reg(s_0, \pi).
\end{equation}
\vspace{-3.5mm}
 \end{problem}

\begin{proposition}
    \label{prop:equivalence}
    If uncertainties are independent per Def.~\ref{def:independence} then Problem~\ref{prob:1} is equivalent to  Problem~\ref{prob:creg_n1}.
\end{proposition}

\textit{Proof sketch}: For independent uncertainty sets, an adversary which chooses one set of parameters to be applied for the entire game is equivalent to an adversary which may change the parameters each step according to a stationary policy. \qedsymbol


Intuitively, this is because for any independent uncertainty set, fixing the parameters applied at one state-action pair does not restrict the set of parameters choices available at other state-action pairs.
For problems of this form, deterministic stationary policies suffice~\cite{iyengar2005robust}.

Problem~\ref{prob:creg_n1} can be solved by applying minimax VI to the regret Bellman equation in Proposition~\ref{prop:creg}.
In the next section, we present Alg.~\ref{alg:min_creg} which solves a generalisation of Problem~\ref{prob:creg_n1}.
The generalisation optimises minimax regret against an adversary, $\sigma^n$, that may change the parameters every $n$ steps.
To solve Problem~\ref{prob:creg_n1}, we apply Alg.~\ref{alg:min_creg} with $n=1$. 
Proposition~\ref{prop:equivalence} shows that this optimises minimax regret exactly for UMDPs with independent uncertainty sets.\footnote[1]{Following publication, the authors realised that this claim is incorrect. Please see the corrigendum.} 

\vspace{-1.5mm}
\subsection{Approx. Solutions for Dependent Uncertainties}
\vspace{-0.5mm}
For UMDPs with dependent uncertainties, optimising minimax regret exactly is intractable~\cite{ahmed2017sampling}.
A possible approach is to over-approximate the uncertainty by assuming independent uncertainties, and solve Problem~\ref{prob:creg_n1}.
However, this gives too much power to the adversary, allowing parameters from different samples to be realised within the same game.
Thus, the minimax regret computed under this assumption is an over-approximation, and the resulting policy may be overly conservative.
In this section, we propose a generalisation of Problem~\ref{prob:creg_n1}  as a way to alleviate this issue.
We start by  bounding the maximum possible error of the over-approximation associated with solving~Problem~\ref{prob:creg_n1} for UMDPs with dependent uncertainties.

\begin{proposition}
	\label{prop:bound_n1}
	If the expected number of steps for $\pi$ to reach $s_g\in G$ is at most $H$ for any adversary:
	\vspace{-2mm}
	\small
	\begin{multline}
	0 \leq \max_{\sigma^{1}} reg(s_0, \pi) - \max_{\xi_q \in \xi} reg_q(s_0, \pi) \vspace{-1mm}\\ 
	\leq (\delta_C + 2\delta_{V^*}  + 2\delta_T C_{max}H )H, \ \ \textnormal{where}
	\end{multline}
	\vspace{-3.8mm}
	\scriptsize
	\begin{align*}
	|\bar{C}_i(s, a) - \bar{C}_j(s, a)| & \leq \delta_C  & \forall s\in S, a \in A, \xi_i\in \xi, \xi_j \in \xi \\
	{\textstyle\sum}_{s'}|T_i(s, a, s') - T_j(s, a, s')| & \leq 2\delta_T & \forall s\in S, a \in A, \xi_i\in \xi, \xi_j \in \xi \\
	|V_i(s, \pi^*) - V_j(s, \pi^*)| & \leq \delta_{V^*} & \forall s\in S, \xi_i\in \xi, \xi_j \in \xi \\
	\bar{C}_i(s, a) & \leq C_{max} & \forall s\in S, a \in A, \xi_i\in \xi
	\end{align*}
	\normalsize
\end{proposition}

\vspace{-1mm}
\subsubsection{$n$-Step Options}
\vspace{-0.5mm}
Prop.~\ref{prop:bound_n1} shows that decoupling the uncertainties at every step over-approximates the maximum regret. We now introduce an approximation which is more accurate, but requires increased computation. Our approach is to approximate dependent uncertainties by decoupling the uncertainty at only every $n$ steps. This results in Problem~\ref{prob:creg_n}, a generalisation of Problem~\ref{prob:creg_n1} where the agent chooses a policy to execute for $n$ steps, and the adversary, $\sigma^n$, reacts by choosing the MDP sample to be applied for that $n$ steps to maximise the regret. After executing $n$ steps, the game transitions to a new state and the process repeats. Increasing $n$ weakens the adversary by capturing dependence over each $n$ step sequence. As $n \rightarrow \infty$ we recover the original minimax regret definition (Problem~\ref{prob:1}).

 \vspace{-1mm}
 \begin{problem}
 	\label{prob:creg_n}
 	Find the minimax regret policy in the stochastic game defined by
 	 \vspace{-1.5mm}
 	\begin{equation}
 	\label{eq:creg_decoupled}
 	\pi_{reg}^n = \argmin_{\pi \in \Pi} \max_{\sigma^n} reg(s_0, \pi).
 	\end{equation}
 \vspace{-3mm}
 \end{problem}

In the remainder of this section, we present our approach to solving Problem~\ref{prob:creg_n}. We start by defining $n$-step options, an adaption of options~\cite{sutton1999between}.

\begin{definition}
	An $n$-step option is a tuple $o=(\bar{s}, \pi^o, G^o, n)$, where $\bar{s}$ is the initiation state where the option may be selected, $\pi^o$ is a policy, $G^o$ is a set of goal states, and $n$ is the number of steps.
\end{definition}
\vspace{-1mm}
If an $n$-step option is executed at $\bar{s}$, the policy $\pi^o$ is executed until one of two conditions is reached: either $n$ steps pass, or a goal state $s_g \in G^o$ is reached. Hereafter, we assume that the goal states for all $n$-step options coincide with the goal states for the UMDP, $G^o$\hspace{3pt}=\hspace{3pt}$G$.
The probability of reaching $s'$ after executing option $o$ in $\bar{s}$ is denoted by $\Pr(s' | \bar{s}, o)$.

We are now ready to define the $n$-step option MDP ($n$-MDP).
The $n$-MDP is equivalent to the original MDP, except that we reason over options which represent policies executed for $n$ steps in the original MDP.
 Additionally, the cost for applying option $o$ at $s$ in the $n$-MDP is equal to the regret attributed to applying $\pi^o$ at $s$ in the original MDP for $n$ steps according to the regret Bellman equation in Proposition~\ref{prop:creg}.

\begin{definition}
\label{def:nmdp}
An $n$-step option MDP (n-MDP) $\mathcal{M}^n$, of original SSP MDP $\mathcal{M}$, is defined by the tuple ($S, s_0, O, C^o, T^o, G$). $S$, $s_0$, and $G$ are the same as in the original MDP. $O$ is the set of possible $n$-step options. ${T^o: S \times O \times S \rightarrow [0, 1]}$ is the transition function for applying options, where $T^o(s, o, s') = \Pr(s' | s, o)$. The cost function, $C^o: S \times O \rightarrow {\rm I\!R}$ is defined as:

\small
\vspace{-1mm}
\begin{equation}
\label{eq:cost_fn}
    C^o(s, o) = V^n(s, \pi^o) + {\textstyle\sum}_{s' \in S}T^o(s,o, s') \cdot V(s', \pi^*) - V(s, \pi^*),
\end{equation}
\normalsize
\noindent where $V^n(s, \pi^o)$ is the expected value for applying $\pi^o$ for $n$ steps starting in $s$.
\end{definition}
\vspace{-0.5mm}

The policy that selects options for the $n$-MDP is denoted $\pi^n$.
We can convert a UMDP to a corresponding $n$-UMDP by converting each MDP sample in the UMDP to an $n$-MDP.

\begin{proposition}
	\label{prop:robust_equiv}
	 Problem~\ref{prob:creg_n} is equivalent to finding the robust policy (Eq.~\ref{eq:robust}) for the $n$-UMDP.
\end{proposition}

\textit{Proof sketch:} The regret Bellman equation in {Proposition~\ref{prop:creg}} can equivalently be written for an $n$-MDP as

\footnotesize
\vspace{-1mm}
\begin{multline}
    \label{eq:creg_as_cost}
 reg(s, \pi^n) = \sum_{o \in O} \pi^n(s, o)  [C^o(s,o)  \\
  + \sum_{s' \in S}T^{o}(s, o, s')  reg(s', \pi^n) ].
 \raisetag{0.0\normalbaselineskip}
\end{multline}
\vspace{-3mm}
\normalsize

\noindent This is an MDP Bellman equation using the cost and transition functions for the $n$-MDP. Therefore, finding the minimax regret according to Problem~\ref{prob:creg_n} is equivalent to finding the minimax expected cost for the $n$-UMDP. This is optimised by the robust policy for the $n$-UMDP.$~\qedsymbol$

\vspace{-0.5mm}
 \subsubsection{Minimax Value Iteration} Prop.~\ref{prop:robust_equiv} means we can use minimax VI on the $n$-MDP to solve Problem~\ref{prob:creg_n} via the recursion

\vspace{-2mm}
\small
\begin{multline}
\label{eq:minimax_cr}
\hspace{-12pt}
reg(s, \pi^n) = \min_{o \in O} \max_{\xi_q \in \xi}[C^o_q(s,o)  +  {\textstyle\sum}_{s' \in S}T^{o}_q(s, o, s') reg(s', \pi^n) ].
 \hspace{-12pt}
\raisetag{0.4\normalbaselineskip}
\end{multline}
\normalsize

\vspace{-2mm}
The results for minimax VI apply~\cite{iyengar2005robust, nilim2005robust}, and therefore the optimal policy is stationary and deterministically chooses options, $\pi^n : S \times O \rightarrow \{0, 1\}$.
 To guarantee convergence, we apply a perturbation by adding a small scalar $\kappa > 0$ to the cost in Eq.~\ref{eq:cost_fn}. It can be shown that in the limit as $\kappa \rightarrow 0^+$, the resulting value function approaches the exact solution~\cite{bertsekas2018abstract}. 
 
 Algorithm~\ref{alg:min_creg} presents pseudocode for the minimax VI algorithm. In Line~\ref{alg:l1}, we start by computing the optimal value in each of the MDP samples using standard VI. This is necessary to compute the contribution to the regret of any action according to Proposition~\ref{prop:creg}. Line~\ref{alg:l2} initialises the values for the minimax regret of the policy to zero at all states. In Lines~\ref{alg:l5}-\ref{alg:l10} we sweep through each state until convergence. At each state we update both the minimax regret value and the option chosen by the policy, according to the Bellman backup defined by Eq.~\ref{eq:minimax_cr}. Solving Equation~\ref{eq:minimax_cr} is non-trivial, and we formulate a means to solve it in the following subsection.

 \setlength{\textfloatsep}{4pt} 
\begin{algorithm}[t]
	\footnotesize
	\begin{algorithmic}[1]
		\State compute $V_q(s, \pi^*_q) \ \forall s \in S, \xi_q \in \xi$ \label{alg:l1}
		\State $reg(s, \pi^n) \leftarrow 0,\ \forall s \in S$ \label{alg:l2}
		\Repeat
		\State $\Delta \leftarrow 0$
		\For{$\bar{s} \in S$} \label{alg:l5}
		\State $reg_{old} \leftarrow reg(\bar{s}, \pi^n)$
		\State \label{alg:line1} $reg(\bar{s}, \pi^n) \leftarrow \min_{o \in O} \max_{\xi_q \in \xi}[C^o_q(\bar{s},o)  +  $ \par \hskip\algorithmicindent \hskip\algorithmicindent
		\hskip\algorithmicindent 
		\hskip\algorithmicindent$ \sum_{s'}T^{o}_q(\bar{s}, o, s') \cdot reg(s', \pi^n) ]$ ~~(Eq.~\ref{eq:minimax_cr})\label{alg:l7}
		\State \label{alg:line2} $\pi^{n}(\bar{s}) \leftarrow \argmin_{o \in O} \max_{\xi_q \in \xi}[ C^o_q(\bar{s}, o)  + $ \par 
		\hskip\algorithmicindent \hskip\algorithmicindent \hskip\algorithmicindent
		\hskip\algorithmicindent $ \sum_{s'}T^{o}_q(\bar{s}, o, s') \cdot reg(s', \pi^n) ]$ ~~(Eq.~\ref{eq:minimax_cr}) \label{alg:l8}
		\hspace{-3mm}
		\State $\Delta \leftarrow \max(\Delta, |reg(\bar{s}, \pi^n) - reg_{old}|)$ \vspace{-0.1mm}
		\EndFor \label{alg:l10} \vspace{-0.1mm}
		\Until{$\Delta < \epsilon$} \vspace{-1mm} \label{alg:l11} 
		\caption{Minimax Value Iteration for Minimax Regret Optimisation with $n$-Step Options \label{alg:min_creg}\vspace{-0.4mm}}
	\end{algorithmic}
\end{algorithm}
\normalsize

 Eq.~\ref{eq:upper_rpop} of Prop. ~\ref{prop:bound_genn} establishes that for dependent uncertainties, $\max_{\sigma^{n}} reg(s_0, \pi)$ is an upper bound on the maximum regret for the policy for any $n$. Eq.~\ref{eq:factor_increase} shows that if we increase $n$ by a factor $k$ and optimise the policy using Algorithm~\ref{alg:min_creg}, we are guaranteed to equal or decrease this upper bound.
  Our experiments demonstrate that in practice increasing $n$ improves performance substantially.
  
  \vspace{-0.5mm}
\begin{proposition}
	For dependent uncertainty sets, 
	\vspace{-1mm}
\label{prop:bound_genn}
	\begin{equation}
		\label{eq:upper_rpop}
		\max_{\sigma^{n}} reg(s_0, \pi) - \max_{\xi_q \in \xi} reg_q(s_0, \pi) \geq 0 \hspace{10pt} \forall\ n \in \mathbb{N},
	\end{equation}
	\vspace{-4.5mm}
    \begin{equation}
    	\label{eq:factor_increase}
    \min_{\pi^n} \max_{\sigma^{n}} reg(s_0, \pi^n) \geq \min_{\pi^{kn}} \max_{\sigma^{kn}} reg(s_0, \pi^{kn})\hspace{7pt} \forall\ n, k \in \mathbb{N}.
    \end{equation}
\end{proposition}
\normalsize
\vspace{-2mm}

\subsubsection{Optimising the Option Policies}
\vspace{-0.5mm}
To perform minimax VI in Algorithm~\ref{alg:min_creg}, we repeatedly solve the Bellman equation defined by Eq.~\ref{eq:minimax_cr}. Eq.~\ref{eq:minimax_cr} corresponds to finding an option policy by solving a finite-horizon minimax regret problem with dependent uncertainties.
Because of the dependence over the $n$ steps, the optimal option policy may be  history-dependent~\cite{steimle2018multi,wiesemann2013robust}.
Intuitively, this is because for dependent uncertainty sets, the history may provide information about the uncertainty set at future stages.
To maintain scalability, whilst still incorporating some memory into the option policy, we opt to consider option policies which depend only on the state and time step, $t$.
Therefore, the optimisation problem in Eq.~\ref{eq:minimax_cr} can be written as Table~\ref{tab:optim_prob}.

In Table~\ref{tab:optim_prob}, we optimise $reg(\bar{s}, \pi^n)$, the updated value for the minimax regret at $\bar{s}$. The other optimisation variables are those denoted by $c_q, V_q^n$, and $\pi^o$. 
The optimal value in each sample, $V_q(s, \pi_q^*)$, is precomputed in Line~\ref{alg:l1} of Alg~\ref{alg:min_creg}. 
$reg(s', \pi^n)$ is the current estimate of the minimax regret for $\pi^n$ at each state, and is initialised to zero in Line~\ref{alg:l2} of Alg~\ref{alg:min_creg}.

The constraints in Table~\ref{tab:optim_prob} represent the following.
The set $S^q_{\bar{s}, t}$ contains all the states reachable in exactly $t$ steps, starting from $\bar{s}$, in sample $\xi_q$.
Eq.~\ref{eq:worst_case} corresponds to the regret Bellman equation for options in Eq.~\ref{eq:cost_fn}-\ref{eq:creg_as_cost}, where the inequality over all $\xi_q$ enforces minimising the maximum regret.
The variables denoted $c_q(s, t)$ represent the expected cumulative part of the minimax regret in Eq.~\ref{eq:cost_fn}-\ref{eq:creg_as_cost} resulting from the expected state distribution after applying $\pi^o$ in sample $\xi_q$ for the horizon of $n$ steps.
Constraint equations~\ref{eq:creg_start}-\ref{eq:creg_end} propagate these $c_q$ values over the $n$ step horizon.
The variables denoted $V^n_q(s, t)$ represent the expected value over the $n$-step horizon of $\pi^o$ at time $t$ in sample $\xi_q$, and the computation of the expected value is enforced by the constraints in Eq.~\ref{eq:value_start}-\ref{eq:value_end}.
%

In the supplementary material, we provide linearisations for the nonlinear constraints in Eq.~\ref{eq:creg_end} and~\ref{eq:value_end}. We consider both deterministic and stochastic option policies, as due to the dependent uncertainties the optimal option policy may be stochastic~\cite{wiesemann2013robust}. The solution is exact for deterministic policies, and for stochastic policies a piecewise linear approximation is required.

\label{sec:inner_problem}
\begin{table}[t!]
\footnotesize
\begin{flalign*}
    \underset{}{\textbf{min }}   reg(\bar{s}, \pi^n) \textbf{ s. t. } &&
\end{flalign*}
\vspace{-4mm}
\begin{equation}
\resizebox{0.88\hsize}{!}{%
$reg(\bar{s}, \pi^n) \geq V^n_q(\bar{s},t) - V_q(\bar{s},\pi^*_q) + c_q(\bar{s}, t),  \hspace{5pt}\forall \xi_q,  t = 0 \label{eq:worst_case}$
}
\end{equation}
\vspace{-7mm}
\begin{multline}
c_q(s, a, t) = {\textstyle\sum}_{s'} T_q(s, a, s') \cdot [reg(s', \pi^n) + V_q(s', \pi^*_q)], \\ \forall s\in S^q_{\bar{s}, t}, a, \xi_q, t = n-1 \label{eq:creg_start}
\end{multline}
\vspace{-7mm}
\begin{multline}
c_q(s, a, t)  = {\textstyle\sum}_{s'} T_q(s, a, s') \cdot c_q(s', t+1), \\ \forall s\in  S^q_{\bar{s}, t}, a, \xi_q, t < n-1
\end{multline}
\vspace{-7mm}
\begin{multline}
c_q(s, t) = {\textstyle\sum}_a \pi^o(s,a) \cdot c_q(s,a,t), \\ \forall s \in S^q_{\bar{s}, t}, \xi_q, t \leq n-1 \label{eq:creg_end}
\end{multline}
\vspace{-7mm}
\begin{multline}
\resizebox{0.85\hsize}{!}{%
$V_q^n(s,a,t) = \bar{C}_q (s, a), \hspace{15pt} \forall s \in S^q_{\bar{s}, t}, a, \xi_q, t = n-1 
\label{eq:value_start}$
}
\end{multline}
\vspace{-7mm}
\begin{multline}
V_q^n(s,a,t) = \bar{C}_q (s, a) + {\textstyle\sum}_{s'}T_q(s, a, s') \cdot V_q^n(s',t+1),  \\
\forall s \in  S^q_{\bar{s}, t}, a, \xi_q, t < n-1
\end{multline}
\vspace{-7mm}
\begin{multline}
V^n_q(s,t) = {\textstyle\sum}_a \pi^o(s,a) \cdot V^n_q(s,a,t), \\  \forall s \in S^q_{\bar{s}, t}, \xi_q, t \leq n-1 
\label{eq:value_end}
\end{multline}
\vspace{-6.5mm}
\caption{Formulation of the optimisation problem over option policies in Equation~\ref{eq:minimax_cr}.}
\label{tab:optim_prob}
\vspace{-0.5mm}
\end{table}
\normalsize

\vspace{-0.5mm}
\subsection{Discussion}
\vspace{-0.5mm}
\paragraph{Complexity}

For SSP MDPs with positive costs, the number of iterations required for VI to converge within residual $\epsilon$ is bounded by $\mathcal{O}(||V^*||^2|S|^2/g^2 + (\log ||V^*|| + \log \epsilon) ||V^*|| \hskip 1pt |S| / g)$, where $g$ is the minimum cost, $V^*$ is the optimal value, and $||x||$ is the $L_\infty$ norm of $x$~\cite{bonet2007speed}.
In our problem, the minimum cost is $\kappa$. During each VI sweep, we solve Eq.~\ref{eq:minimax_cr} $|S|$ times by optimising Table~\ref{tab:optim_prob}.
Therefore, Table~\ref{tab:optim_prob} must be solved  $\mathcal{O}(||reg^*||^2|S|^3/\kappa^2 + (\log ||reg^*|| + \log \epsilon) ||reg^*|| \hskip 1pt |S|^2 / \kappa)$ times, where $reg^*$ is the optimal minimax regret for Problem~\ref{prob:creg_n}.
To assess the complexity of optimising the model in Table~\ref{tab:optim_prob}, assume the MDP branching factor is $k$.
Then the number reachable states in $n$ steps is $\mathcal{O}(k^n)$, and the size of the model is $\mathcal{O}(|A||\xi|nk^n)$.
MILP solution time is exponential, and therefore complexity is $\mathcal{O}(exp(|A||\xi|nk^n))$.
Crucially, $|S|$ is not in the exponential.

\vspace{-3.3mm}
\paragraph{Sampling}
\label{sec:uncert}
This approach requires a finite set of samples, yet for some problems there are infinite possible MDP instantiations. To optimise worst-case regret, we need samples which provide adequate coverage over possible MDP instantiations so that the resulting worst-case solution generalises to all possible MDPs.
Where necessary, we use the sampling approach proposed for this purpose in~\cite{ahmed2013regret}.

\vspace{-3.3mm}
\paragraph{Action Pruning}
\label{sec:pruning}
To reduce the size of the problem in Table~\ref{tab:optim_prob}, we can prune out actions that are unlikely to be used by the minimax regret policy.
We propose the following pruning method, analagous to the approach proposed by~\citet{lacerda2017multi}.
The policy with optimal expected cost, $\pi_q^*$, is computed for each $\xi_q \in \xi$ to create the set of optimal policies $\Pi^*_\xi = \{\pi_1^*, \pi_2^*,\ldots, \pi_{|\xi|}^* \}$.
We build the pruned UMDP by removing transitions $T_q (s, a, s')$ for all $\xi_q$, for which $a$ is not in $\pi(s)$ for some policy $\pi \in \Pi^*_\xi$.
The intuition is that actions which are directed towards the goal are likely to be included in at least one of the optimal policies.
Therefore, this pruning process removes actions which are not directed towards the goal, and are unlikely to be included in a policy with low maximum regret.

\vspace{-2mm}
\section{Evaluation}
\vspace{-0.5mm}
We evaluate the following approaches on three domains with dependent uncertainties:  
\vspace{-1.2mm}
\begin{itemize}[leftmargin=*,wide=0pt]
	\itemsep-1.7pt 
	\item \textit{reg}: the approach presented in this paper. 
	\item \textit{cemr}: the state of the art approach from~\cite{ahmed2013regret, ahmed2017sampling} which we have extended for $n$-step options. 
	\item \textit{robust}: the standard robust dynamic programming solution in Eq.~\ref{eq:robust}~\cite{iyengar2005robust, nilim2005robust}.
	\item \textit{MILP}: the optimal stationary deterministic minimax regret policy computed using the MILP in~\cite{ahmed2013regret}. We do not compare against the stochastic version as we found it did not scale beyond very small problems.
	\item \textit{Averaged MDP}: this benchmark averages the cost and transition parameters across the MDP samples and computes the optimal policy in the resulting MDP~\cite{adulyasak2015solving, chen2012tractable,delage2010percentile}.
	\item \textit{Best MDP Policy}: computes the optimal policy set, $\Pi^*_\xi$. For each policy, $\pi \in \Pi^*_\xi$, the maximum regret is evaluated for all $\xi_q \in \xi$. The policy with lowest maximum regret is selected. 
\end{itemize}

We found that action pruning reduced computation time for \textit{reg}, \textit{cemr}, and \textit{MILP} without significantly harming performance. Therefore we only present results for these approaches with pruning. We write $(s)$ for stochastic $(d)$ for deterministic option policies. MILPs are solved using Gurobi, and all other processing is performed in Python. Computation times are reported for a 3.2 GHz Intel i7 processor. For further details on the experimental domains see~\citet{rigter2020minimax}.

\vspace{-4mm}
\paragraph{Medical Decision Making Domain}
We test on the medical domain from~\citet{sharma2019robust}.
The state, $(h, d)  \in S$ comprises of two factors: the health of the patient, $h \in \{0, \ldots, 19\}$, and the day $d \in \{0, \ldots, 6\}$.
At any state, one of three actions can be applied, each representing different treatments.
In each MDP sample the transition probabilities for each treatment differ, corresponding to different responses by patients with different underlying conditions.
The health of the patient on the final day determines the cost received.

\vspace{-4mm}
\paragraph{Disaster Rescue Domain}
We adapt this domain from the UMDP literature~\cite{adulyasak2015solving, ahmed2013regret, ahmed2017sampling} to SSP MDPs. 
An agent navigates an 8-connected grid which contains swamps and obstacles by choosing from 8 actions.
Nominally, for each action the agent transitions to the corresponding target square with $p=0.8$, and to the two adjacent squares with $p=0.1$ each. 
If the target, or adjacent squares are obstacles, the agent transitions to that square with probability 0.05. 
If the square is a swamp, the cost is sampled uniformly in ${[1, 2]}$. The cost for entering any other state is 0.5. 
The agent does not know the exact locations of swamps and obstacles, and instead knows regions where they may be located.
To construct a sample, a swamp and obstacle is sampled uniformly from each swamp and obstacle region respectively.
Fig.~\ref{fig:disaster_rescue} (left) illustrates swamp and obstacle regions for a particular UMDP. 
Fig.~\ref{fig:disaster_rescue} (right) illustrates a possible sample corresponding to the same UMDP.

\vspace{-4mm}
\paragraph{Underwater Glider Domain}
An underwater glider navigates to a goal location subject to uncertain ocean currents from real-world forecasts. For each UMDP, we sample a region of the Norwegian sea, and sample the start and goal location. The mission will be executed between 6am and 6pm, but the exact time is unknown. As such, the navigation policy must perform well during all ocean conditions throughout the day. We construct 12 MDP samples, corresponding to the ocean current forecast at each hourly interval. Each state is a grid cell with 500m side length. There are 12 actions corresponding to heading directions. The cost for entering each state is sampled in [0.8, 1]. An additional cost of 3 is added for entering a state where the water depth is < 260m and current is > 0.12m/s, penalising operation in shallow water with strong currents. The forecast used was for May 1st 2020 and is available online at {\url{https://marine.copernicus.eu/}}.

\setlength{\abovecaptionskip}{5pt}
\begin{figure*}[t]
	\centering
	\begin{minipage}[b]{0.42\linewidth}
		\vspace{0pt}
		\centering
		\includegraphics[width=0.9\linewidth]{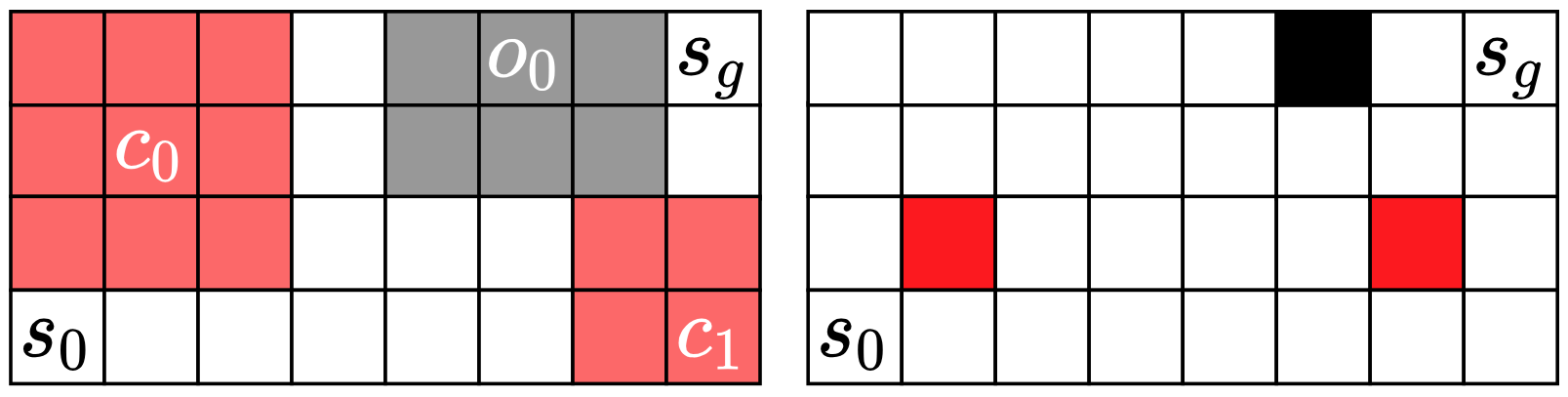}
		\caption{Illustration of an example UMDP in disaster rescue. Left: shaded regions indicate possible swamp and obstacle locations. Right: possible UMDP sample.}
		\label{fig:disaster_rescue}
	\end{minipage}
	\hfill
	\begin{minipage}[b]{0.56\linewidth}
		\vspace{0pt}
		\centering
		\includegraphics[width=0.47\linewidth]{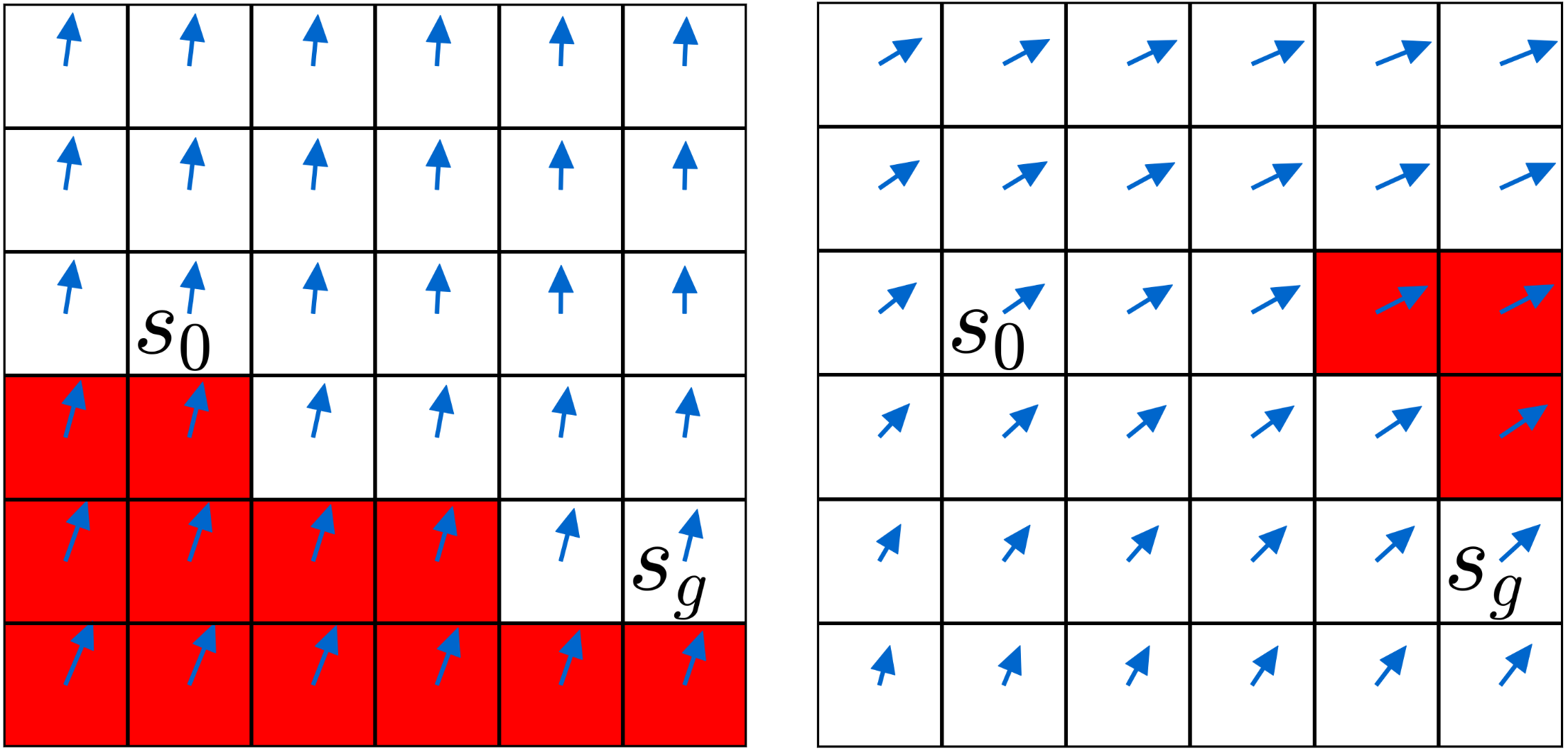}
		\caption{Illustration of two UMDP samples in glider domain. Arrows indicate ocean currents. Red squares indicate states with additional cost.}
		\label{fig:underwater_glider}
	\end{minipage}\\
	\vspace{-2mm}
	\centering
	\begin{minipage}[t][][b]{\textwidth}
		\centering
		\includegraphics[width=1.\linewidth]{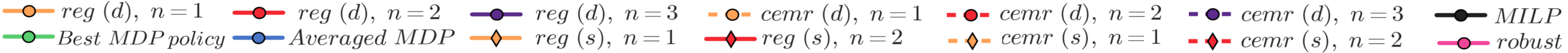}
	\end{minipage}\\
	\begin{minipage}[t][][b]{.32\textwidth}
		\centering
		\includegraphics[width=0.8\linewidth]{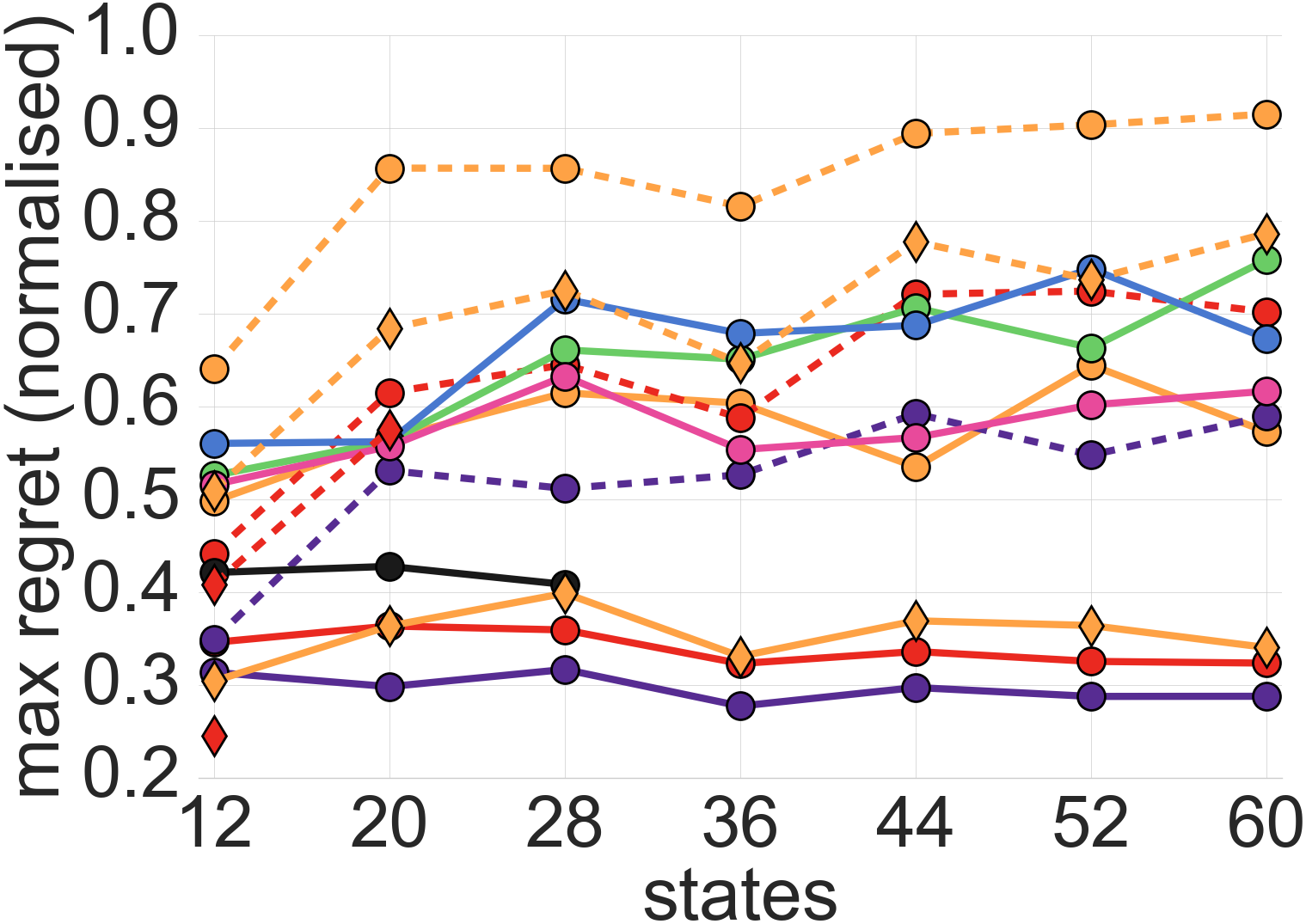}
	\end{minipage}%
	\hfill
	\begin{minipage}[t][][b]{.28\textwidth}
		\centering
		\includegraphics[width=0.94\linewidth]{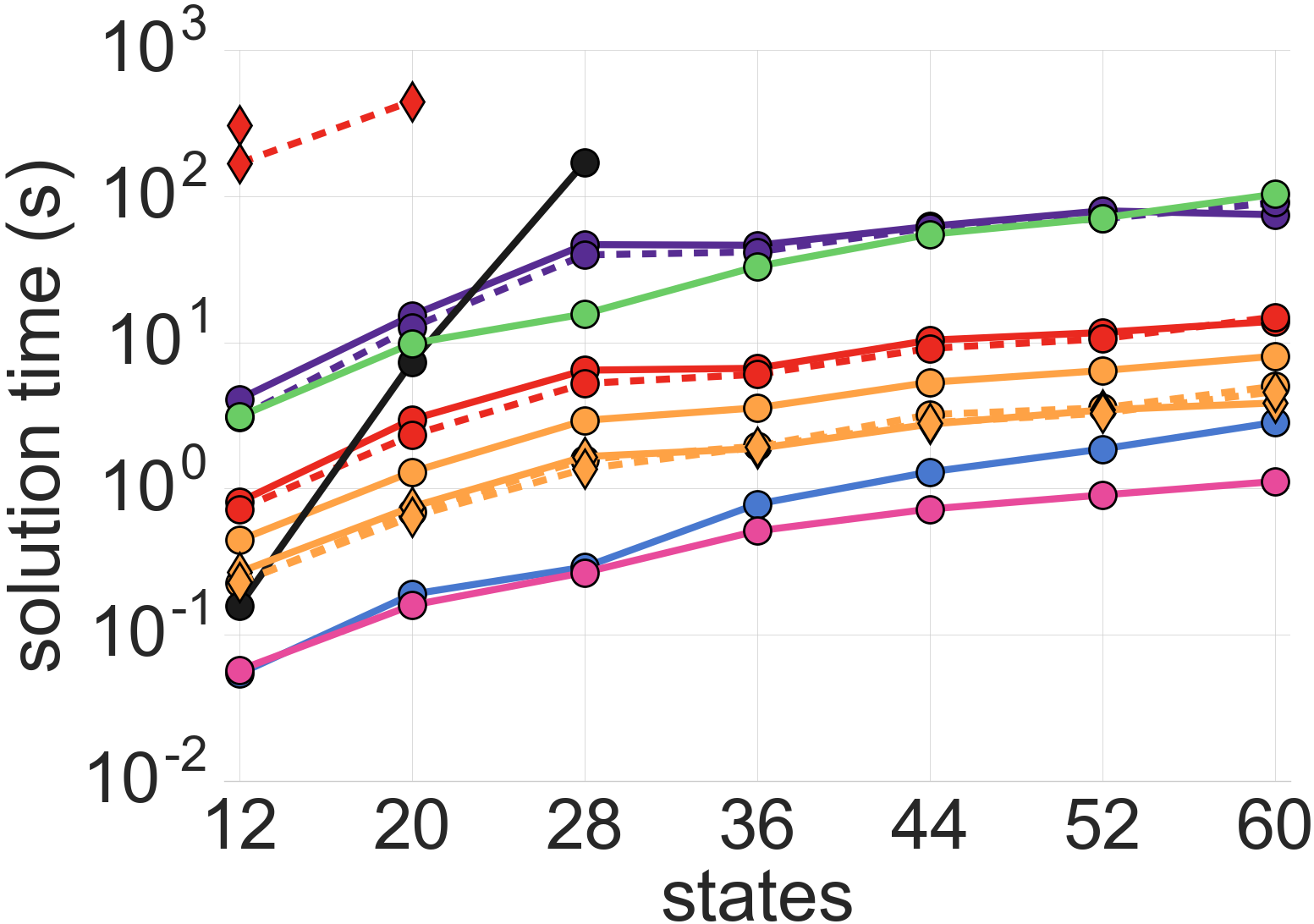}
	\end{minipage}
	\hfill
	\begin{minipage}[t][][b]{.34\textwidth}
		\centering
		\includegraphics[width=0.77\linewidth]{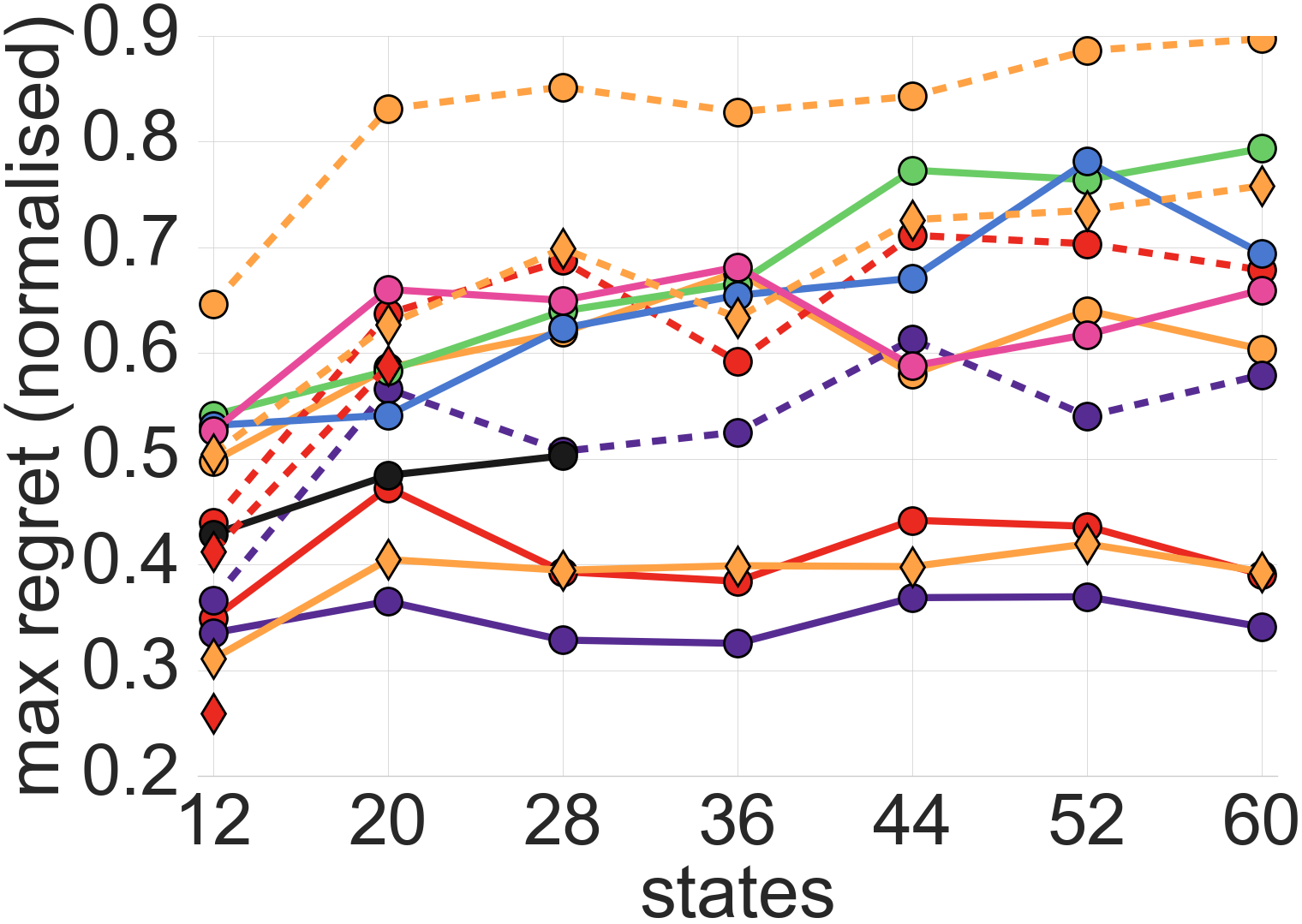}
	\end{minipage}\\
	\begin{minipage}[t][][b]{.32\textwidth}
		\centering
		\includegraphics[width=0.8\linewidth]{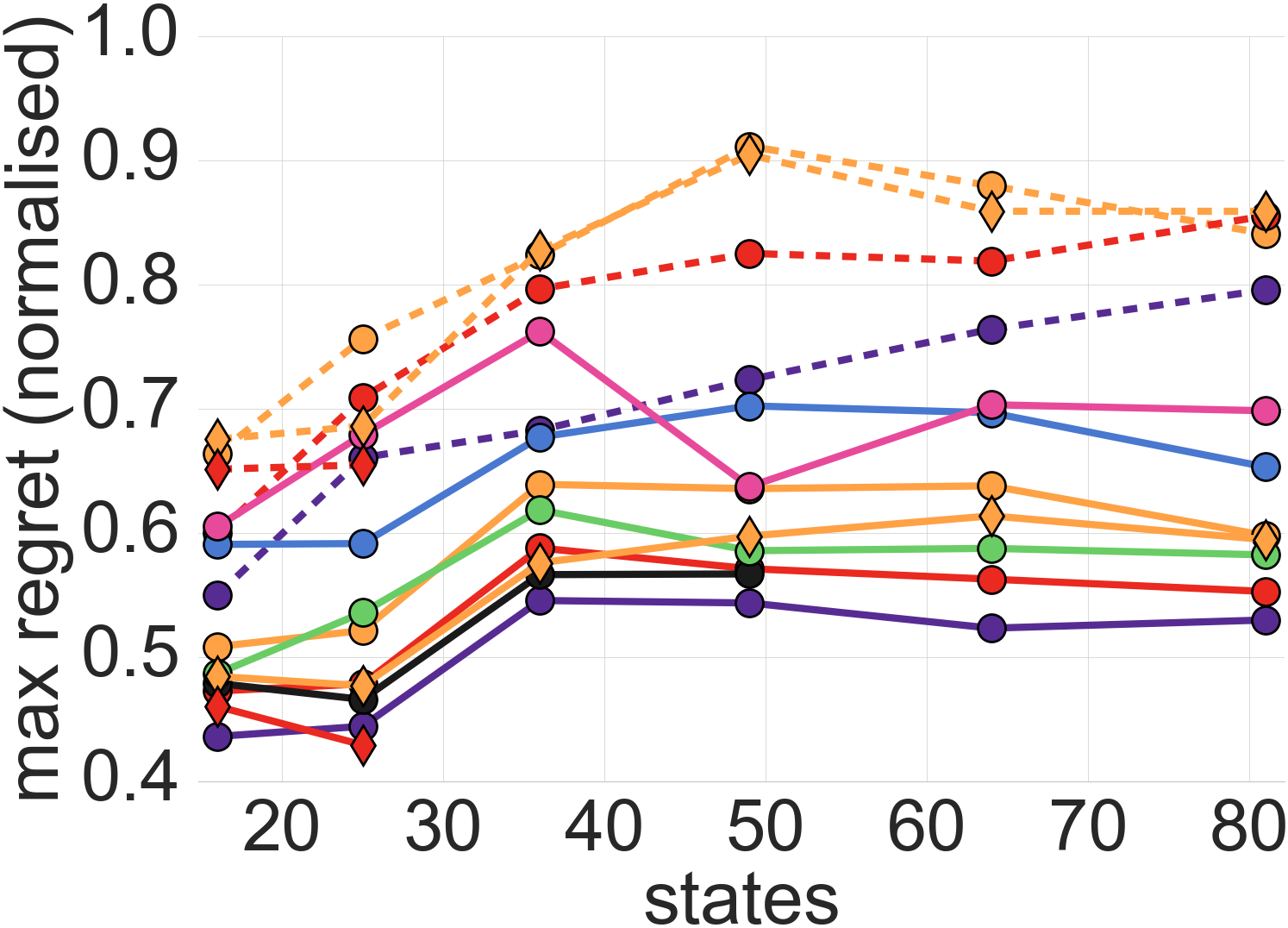}
		\caption{Mean normalised max regret. Top: disaster rescue, bottom: glider.\label{fig:soln_regret}}
	\end{minipage}%
	\hfill
	\begin{minipage}[t][][b]{.28\textwidth}
		\centering
		\includegraphics[width=0.94\linewidth]{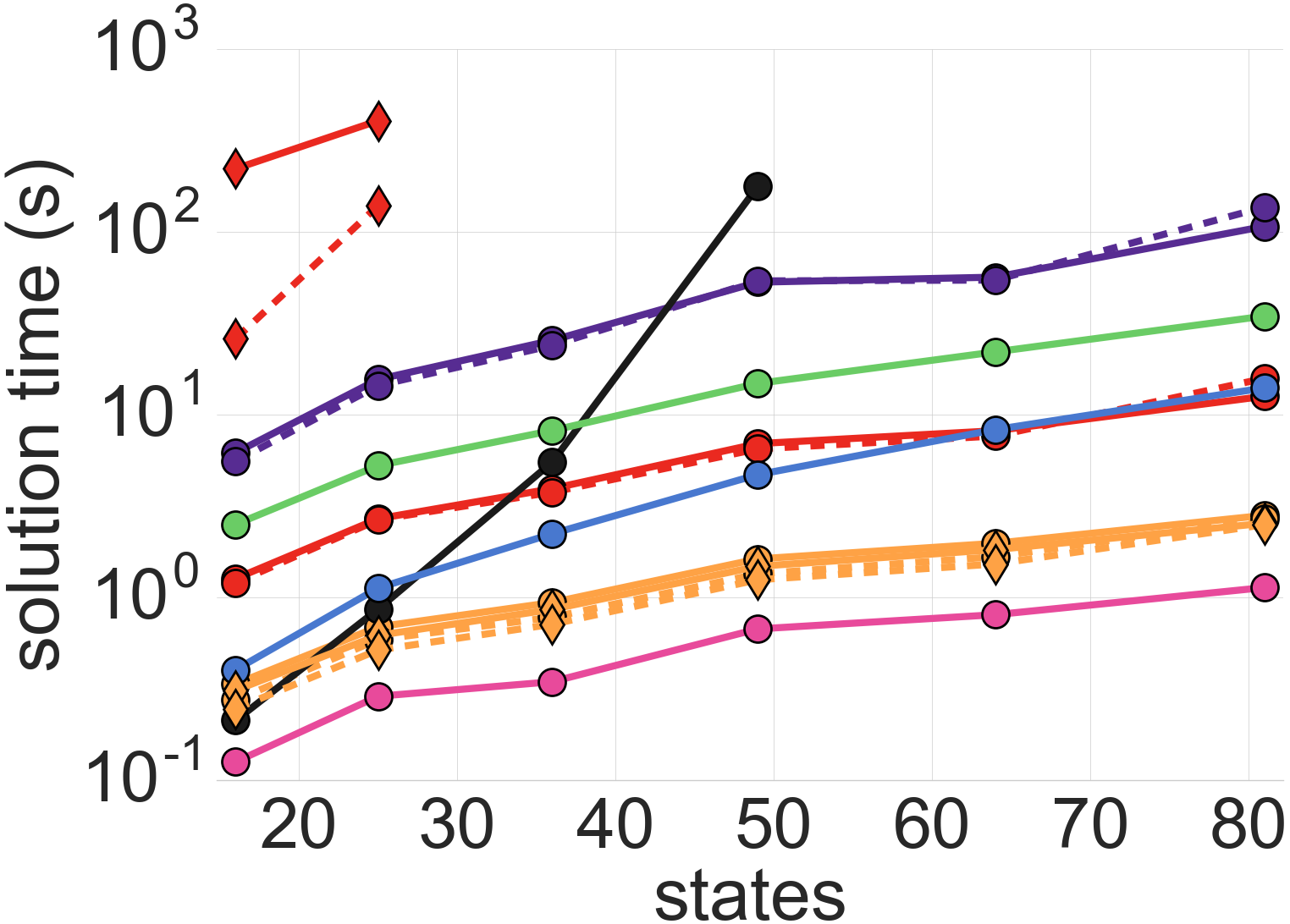}
		\caption{Mean solution times. Top: disaster rescue, bottom: glider.\label{fig:times}}
	\end{minipage}
	\hfill
	\begin{minipage}[t][][b]{.34\textwidth}
		\centering
		\includegraphics[width=0.78\linewidth]{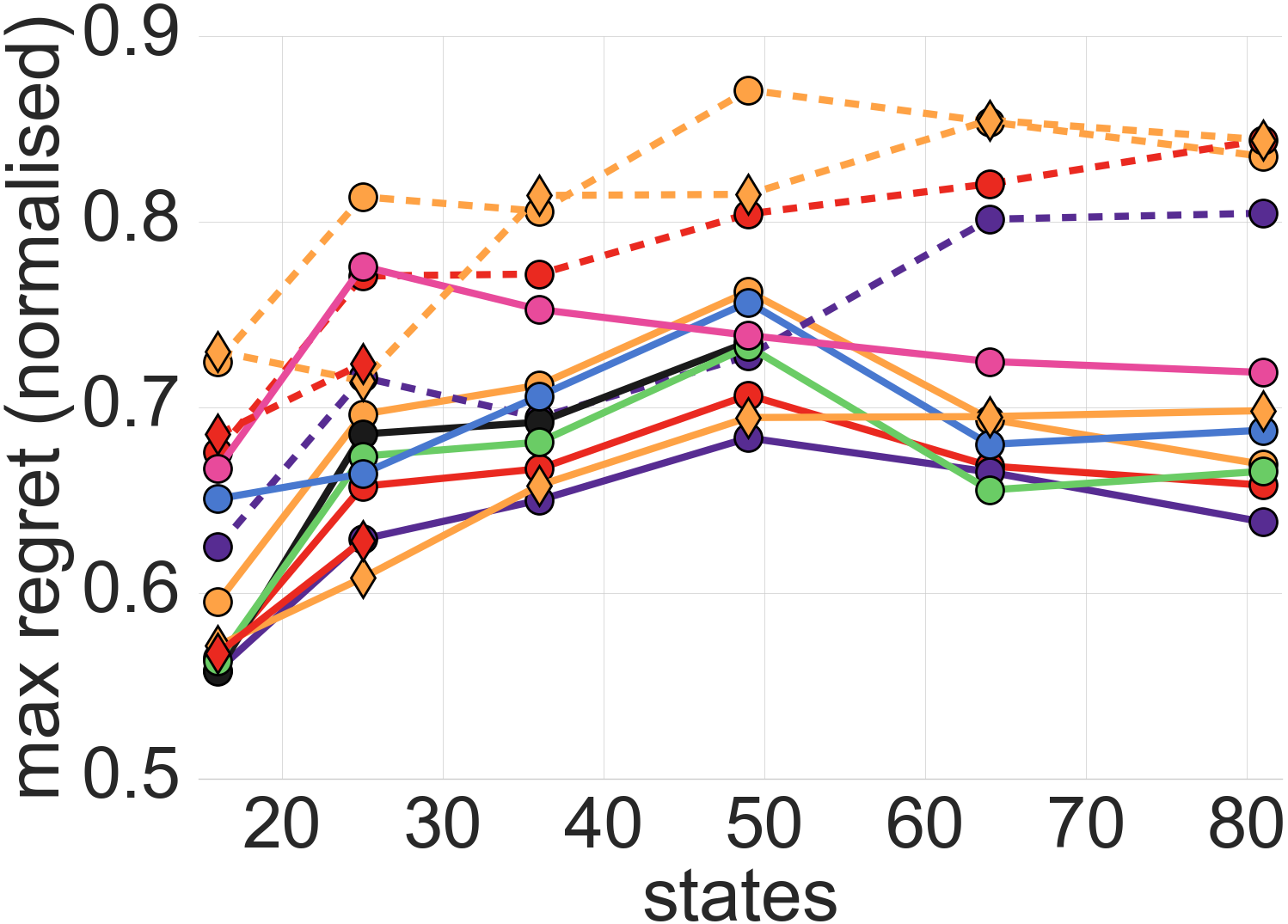}
		\caption{Mean normalised max regret on test set. Top: disaster rescue, bottom: glider.\label{fig:eval_regret}}
	\end{minipage}\\
	\vspace{1mm}
	\setlength{\abovecaptionskip}{10pt}
	\footnotesize
	\begin{minipage}[t][][b]{\textwidth}
	\resizebox{\textwidth}{!}{%
	\renewcommand{\arraystretch}{1.07} 
	\setlength\tabcolsep{2pt}
	\begin{tabular}{| l | c | c | c | c | c | c | c | c |c | c| c| }
		\hline
		\footnotesize{Method}                 &  \includegraphics[scale=0.22]{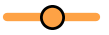} & \includegraphics[scale=0.22]{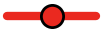} & \includegraphics[scale=0.22]{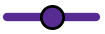} & \includegraphics[scale=0.22]{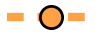} & \includegraphics[scale=0.22]{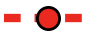} & \includegraphics[scale=0.22]{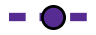} & \includegraphics[scale=0.22]{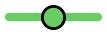}&
		\includegraphics[scale=0.22]{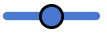}&
		\includegraphics[scale=0.22]{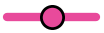}&
		\includegraphics[scale=0.22]{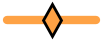}&
		\includegraphics[scale=0.15]{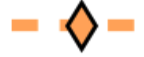}  \\
		\hline
		\footnotesize{max reg}                & 0.596; 0.22 & 0.538; 0.18& $\mathbf{0.497}$; 0.16&0.906; 0.11&0.885; 0.13&0.880; 0.12&0.557; 0.20&0.596; 0.22&0.641; 0.23&0.529; 0.15&0.846; 0.12\\
		\hline
		\footnotesize{time (s)}                & 5.03; 0.40& 21.1; 2.3& 77.6; 18 &4.24; 0.33&20.0; 2.4&66.9; 16&103; 7.9& 3.75; 0.08&$\mathbf{0.391}$; 0.03&4.64; 0.35&3.96; 0.31\\
		\hline
		\footnotesize{max reg test set} 	& 0.674; 0.19 &  0.636; 0.18 & 0.625; 0.18 & 0.870; 0.13 &0.855; 0.14 &0.852; 0.13&0.706; 0.20& 0.677; 0.20 &0.715; 0.19 & $\mathbf{0.574}$; 0.12&0.814;0.13 \\
		\hline
	\end{tabular}
	}
	\vspace{-2mm}
	\normalsize
	\captionof{table}{Mean normalised maximum regret in the medical domain, averaged over 250 UMDPs. Methods which did not find a solution within an average time of 600s are not included in the table. Format: mean; standard deviation. \label{tab:medical_res}}
	\end{minipage}\\
\vspace{-5mm}
\end{figure*}

\vspace{-2mm}
\subsection{Experiments}
\label{sec:max_reg}
\vspace{-1mm}
\paragraph{Maximum Regret}
For the medical domain, each method was evaluated for 250 different randomly generated UMDPs. 
For the other two domains, each method was evaluated for a range of problem sizes, and
each problem size was repeated for 25 different randomly generated UMDPs. 
For each disaster rescue and medical decision making UMDP, $\xi$ consisted of 15 samples selected using the method from~\cite{ahmed2013regret, ahmed2017sampling}. 
In underwater glider, $\xi$ consisted of the 12 samples corresponding to each hourly weather forecast.
For each method, we include results where the average computation time was <600s. 
For the resulting policies, the maximum regret over all samples in $\xi$ was computed.
Each max regret value was normalised between [0, 1] by dividing by the worst max regret for that UMDP across each of the methods.
The normalised values were then averaged over all 25/250 runs, and displayed in Fig~\ref{fig:soln_regret} and the top row of Table~\ref{tab:medical_res}.

\vspace{-4mm}
\paragraph{Generalisation Evaluation}
For sample-based UMDPs, the policy is computed from a finite set of samples.
To assess generalisation to any possible MDP instantiation, we evaluated the maximum regret on a larger set of 100 random samples.
For disaster rescue and medical decision making, the samples were generated using the procedure outlined.
For underwater glider, we generated more samples by linearly interpolating between the 12 forecasts and adding Gaussian noise to the ocean current values ($\sigma=$ 2\% of the ocean current velocity).
The average maximum regret for this experiment is shown in Fig.~\ref{fig:eval_regret} and the bottom row of Table~\ref{tab:medical_res}.

\vspace{-3.5mm}
\paragraph{Results}
A table of $p$-values is included in the supplementary material which shows that most differences in performance between methods are statistically significant, with the exception of those lines in the figures which overlap.
Fig.~\ref{fig:soln_regret} shows that in general, our approach significantly outperforms the baselines with the exception of \textit{MILP}, which also has good performance.
However, \textit{MILP} scales poorly as indicated by Fig.~\ref{fig:times}, and failed to find solutions in the medical domain within the 600s time limit.
\textit{MILP} finds the optimal deterministic stationary policy considering full dependence between uncertainties.
However, the performance of our approach improves significantly with increasing $n$, and outperforms \textit{MILP} with larger $n$.
This indicates that the limited memory of the non-stationary option policies is crucial for strong performance.
For the same $n$, stochastic option policies improve performance in both domains.
However, the poor scalability of stochastic policies indicates that deterministic options with larger $n$ are preferable. 
Across all domains the current state of the art method, CEMR with $n=1$, performed poorly.
Performance of CEMR improved somewhat by extending it to use our options framework ($n>1$).
The poor performance of CEMR can be attributed to the fact that CEMR
myopically approximates the maximum regret by calculating the performance loss compared to local actions, which may be a poor estimate of the suboptimality over the entire policy.
In contrast, our approach optimises the maximum regret using the recursion given in Prop.~\ref{prop:creg} which computes the contribution of each action to the regret for the policy exactly by comparing against the optimal value function in each sample.

The generalisation results in Fig.~\ref{fig:eval_regret} and Table~\ref{tab:medical_res} show strong performance of our approach for disaster rescue and the medical domain on the larger test set.
In the glider domain, there is more overlap between methods however our approach with larger $n$ still tends to perform the best.
This verifies that in domains with a very large set of possible MDPs, a viable approach is to use our method to find a policy with a smaller set of MDPs and this will generalise well to the larger set.

\vspace{-1.5mm}
\section{Conclusion}
\vspace{-0.5mm}
We have presented an approach for minimax regret optimisation in offline UMDP planning.
Our algorithm solves this problem efficiently and exactly in problems with independent uncertainties.\footnote[1]{Following publication, the authors realised that this claim is incorrect. Please see the corrigendum.}  
To address dependent uncertainties we have proposed using options to capture dependence over sequences of $n$ steps and tradeoff computation and solution quality.
Our results demonstrate that our approach offers state-of-the-art performance.
In future work, we wish to improve the scalability of our approach by using function approximation.

\section{ Acknowledgments}
This work was supported by UK Research and Innovation and EPSRC through the Robotics and Artificial Intelligence for Nuclear (RAIN) research hub [EP/R026084/1], the Clarendon Fund at the University of Oxford, and a gift from Amazon Web Services.

\bibliography{mybib} 

\onecolumn
\newpage

\section{Corrigendum}
Proposition~\ref{prop:equivalence} of the paper is incorrect. The way that the adversary, $\sigma^1$, maximises regret in our approach is not equivalent to the maximising regret over independent uncertainty sets. As a result, our method is approximate for independent uncertainty sets.

\medskip
\noindent 
\textbf{Explanation} \\
\noindent 
The way that our algorithm computes the maximum regret against the adversary, $\max_{\sigma^n} reg(s_0, \pi^n)$, means that for $n = 1$, the maximum regret computed by our algorithm is not equal to the maximum regret over independent uncertainty sets, i.e. $ \max_{\sigma^1} reg(s_0, \pi) \neq \max_{\xi_q \in \xi} reg_q(s, \pi)$ for independent uncertainty set $\xi$. 

Our approach in Algorithm 1 solves Problem~\ref{prob:creg_n} by finding the policy which minimises $\max_{\sigma^n} reg(s_0, \pi^n)$, where this quantity is computed according to:
\begin{multline}
	\label{eq:err28}
	\max_{\sigma^n} reg(s_0, \pi^n) :=  \max_{\xi_q \in \xi} \Bigg\{ \sum_{o \in O} \pi^n(s, o) \cdot \bigg[V^n_q(s, \pi^o) + \sum_{s' \in S}T_q(s, o, s') \cdot V_q(s', \pi^*) - V_q(s, \pi^*)
	+ \\ \max_{\xi_{q'} \in \xi} \sum_{s' \in S}T_q(s, o, s') \cdot reg_{q'}(s', \pi^n) \bigg] \Bigg\}.
\end{multline}

For the case of $n = 1$ we find the policy which minimises $\max_{\sigma^1} reg(s_0, \pi)$ (i.e. Problem~\ref{prob:creg_n1}) which is computed according to:

\begin{multline}
	\label{eq:err27}
	\max_{\sigma^1} reg(s_0, \pi) := \max_{\xi_q \in \xi} \Bigg\{ \sum_{a \in A} \pi(s, a) \cdot \Big[\bar{C}_q(s, a) + \sum_{s' \in S}T_q(s, a, s') \cdot V_{q}(s', \pi^*_q) - \bar{C}_q(s, \pi_q^*(s)) -
	\\
	\sum_{s' \in S}T_q(s, \pi_q^*(s), s') \cdot V_{q}(s', \pi^*_q) + \max_{\xi_{q'} \in \xi} \sum_{s' \in S}T_q(s, a, s')\cdot reg_{q'}(s', \pi) \Big] \Bigg \}
\end{multline}



To illustrate the discrepancy, we compare what is computed by our algorithm in Equation~\ref{eq:err27} to what is derived for the minimax regret using the Bellman equation in Proposition~\ref{prop:creg}. From Proposition~\ref{prop:creg}, the maximum regret over all uncertainty samples, $\xi_q \in \xi$,  can be written as
\begin{multline*}
	\max_{\xi_q \in \xi} reg_q(s, \pi) = \max_{\xi_q \in \xi} \Bigg\{ \sum_{a \in A} \pi(s, a) \cdot  \Big[Q^{gap}_q(s, a) + \sum_{s' \in S}T_q(s, a, s')\cdot reg_q(s', \pi) \Big] \Bigg\} 
	\\
	= \max_{\xi_q \in \xi} \Bigg\{ \sum_{a \in A} \pi(s, a) \cdot \Big[\bar{C}_q(s, a) + \sum_{s' \in S}T_q(s, a, s') \cdot V_q(s', \pi^*_q) - \bar{C}_q(s, \pi_q^*(s)) -
	\\
	\sum_{s' \in S}T_q(s, \pi_q^*(s), s') \cdot V_q(s', \pi^*_q) + \sum_{s' \in S}T_q(s, a, s')\cdot reg_q(s', \pi) \Big] \Bigg \}
\end{multline*}

For independent uncertainty sets we can separately maximise over the member of the uncertainty set to applied at the current step, $\xi_q$, and that to be applied at all subsequent steps, $\xi_{q'}$:

\begin{multline}
	\label{eq:reg_err1}
	\max_{\xi_q \in \xi} reg_q(s, \pi) = \max_{\xi_q, \xi_{q'} \in \xi} \Bigg\{ \sum_{a \in A} \pi(s, a) \cdot \Big[\bar{C}_q(s, a) + \sum_{s' \in S}T_q(s, a, s') \cdot V_{\textcolor{red}{q'}}(s', \pi^*_{q'}) - \bar{C}_q(s, \pi_q^*(s)) -
	\\
	\sum_{s' \in S}T_q(s, \pi_q^*(s), s') \cdot V_{\textcolor{red}{q'}}(s', \pi^*_{q'}) + \sum_{s' \in S}T_q(s, a, s')\cdot reg_{q'}(s', \pi) \Big] \Bigg \}
\end{multline}

Crucially, the value function terms in Equation~\ref{eq:reg_err1}, $V_{q'}$, are computed under the member of the uncertainty set to be applied at all subsequent steps, $\xi_q'$.
However, our approach in Equation~\ref{eq:err27} instead utilises $V_{q}$.
Therefore, intuitively we can think of our approach as optimising the regret accumulated by assuming that the optimal value function at successor states is computed using the worst-case sample for the current state.

As a result of this discrepancy, Proposition~\ref{prop:equivalence} is incorrect and the objective that our approach optimises for independent uncertainty sets is not equal to the maximum regret. Therefore, our algorithm does not provide an exact solution for this class of problems. Theorem 6 of~\cite{ghavamzadeh2016safe} proves that finding the maximum regret of a policy is NP-hard even for independent uncertainty sets, so we cannot expect to find a scalable approach to this problem.

\medskip
\noindent 
\textbf{Impact on the Remainder of the Paper} \\
\noindent 
 Propositions~\ref{prop:bound_n1},~\ref{prop:robust_equiv}, and~\ref{prop:bound_genn} analyse the result of optimising the quantities computed by Equations~\ref{eq:err28} and~\ref{eq:err27} for problems with~\emph{dependent} uncertainty sets.
 The above error only concerns problems with independent uncertainty sets, so does not effect  Propositions~\ref{prop:bound_n1},~\ref{prop:robust_equiv}, and~\ref{prop:bound_genn}. Furthermore, the above error is irrelevant to our experimental results which use domains with dependent uncertainties.


\newpage
\begin{appendices}

\section{Proof of Proposition~\ref{prop:creg}}
\setcounter{proposition}{0}
\begin{proposition}
	(Regret Bellman Equation) The regret for a proper policy, $\pi$, can be computed via the following recursion

	\begin{equation*}
		reg(s, \pi) = \sum_{a \in A} \pi(s, a) \cdot \\ \big[Q^{gap}(s, a) + \sum_{s' \in S}T(s, a, s')\cdot reg(s', \pi) \big], \hspace{5pt} \normalsize{\textit{where}}
	\end{equation*}
	
	\begin{equation*}
		Q^{gap}(s, a) = \big[ \bar{C}(s, a) + \sum_{s' \in S}T(s, a, s') \cdot V(s', \pi^*) \big] - V(s, \pi^*),
		\vspace{-1mm}
	\end{equation*}
and  $reg(s, \pi) = 0,\  \forall s \in G$. 
\end{proposition}

To prove the proposition we show that if we apply the equations in Proposition~\ref{prop:creg} starting from the initial state we recover the definition of the regret for a policy given by Definition~\ref{def:reg_pol}. We start by combining the equations stated in the proposition for the initial state 

\begin{equation}
reg(s_0, \pi) = \sum_{a \in A} \pi(s_0, a) \cdot \\ \Bigg[ \bar{C}(s_0, a) + \sum_{s' \in S}T(s_0, a, s') \cdot V(s', \pi^*)  - V(s_0, \pi^*) + \sum_{s' \in S}T(s_0, a, s')\cdot reg(s', \pi) \Bigg].
\end{equation}

We can move $V(s_0, \pi^*)$ outside of the sum as it does not depend on $a$. We start unrolling the definition by substituting for $reg(s', \pi)$

\begin{multline}
	reg(s_0, \pi) = - V(s_0, \pi^*) + \sum_{a \in A} \pi(s_0, a) \cdot \Bigg[ \bar{C}(s_0, a) + \sum_{s' \in S}T(s_0, a, s') \cdot V(s', \pi^*)   + \\ 
	\sum_{s' \in S}T(s_0, a, s')\cdot \Bigg[ \sum_{a \in A} \pi(s', a) \cdot \Bigg[ \bar{C}(s', a) + \sum_{s'' \in S}T(s', a, s'') \cdot V(s'', \pi^*)  - V(s', \pi^*) + \sum_{s'' \in S}T(s', a, s'')\cdot reg(s'', \pi) \Bigg] \Bigg] \Bigg]. 
	\label{eq:unroll1}
\end{multline}

Again, we can move $V(s', \pi^*)$ outside of the inner sum as it does not depend on $a$. Thus, Equation~\ref{eq:unroll1} can be rewritten as

\begin{multline}
	reg(s_0, \pi) = - V(s_0, \pi^*) + \sum_{a \in A} \pi(s_0, a) \cdot \Bigg[ \bar{C}(s_0, a) + \sum_{s' \in S}T(s_0, a, s') \cdot V(s', \pi^*) +\\ 
	\sum_{s' \in S}T(s_0, a, s')\cdot \Bigg[-V(s', \pi^*) + \sum_{a \in A} \pi(s', a) \cdot \Bigg[ \bar{C}(s', a) + \sum_{s'' \in S}T(s', a, s'') \cdot V(s'', \pi^*)  + \sum_{s'' \in S}T(s', a, s'')\cdot reg(s'', \pi) \Bigg] \Bigg] \Bigg].
	\label{eq:unroll2}
\end{multline}

Cancelling terms, we have

\begin{multline}
	reg(s_0, \pi) = - V(s_0, \pi^*) + \sum_{a \in A} \pi(s_0, a) \cdot \Bigg[ \bar{C}(s_0, a) +  \\ 
	\sum_{s' \in S}T(s_0, a, s')\cdot \Bigg[\sum_{a \in A} \pi(s', a) \cdot \Bigg[ \bar{C}(s', a) + \sum_{s'' \in S}T(s', a, s'') \cdot V(s'', \pi^*)  + \sum_{s'' \in S}T(s', a, s'')\cdot reg(s'', \pi) \Bigg] \Bigg] \Bigg].
\end{multline}

After repeating the above process of unrolling the expression and cancelling terms for $h$ steps we arrive at the following expression

\begin{multline}
	reg(s_0, \pi) = - V(s_0, \pi^*) +
	\sum_{a \in A} \pi(s_0, a) \cdot \bigg[\bar{C}(s_0, a)  +
	\sum_{s' \in S}T(s_0, a, s') \cdot \bigg[\sum_{a \in A}\pi(s', a) \cdot \bigg[\bar{C}(s', a)  + \ldots  \\
	\sum_{s^{h\textnormal{-}1} \in S}T(s^{h\textnormal{-}2}, a, s^{h\textnormal{-}1}) \cdot \bigg[ \sum_{a \in A} \pi(s^{h\textnormal{-}1}, a) \cdot \bigg[ \bar{C}(s^{h\textnormal{-}1}, a) + 
	\sum_{s^{h} \in S}T(s^{h\textnormal{-}1}, a, s^{h}) \cdot V(s^h, \pi^*)
	+  \sum_{s^h \in S}T(s^{h\textnormal{-}1}, a, s^h) \cdot reg(s^h, \pi) \bigg] \bigg] \ldots \bigg] \bigg] \bigg].
\end{multline}

Taking $h \rightarrow \infty$ we have $s^h \in G$ under the definition of a proper policy. Thus, $V(s^h, \pi^*) = 0$ by the definition of goal states in an SSP MDP (Definition~\ref{def:ssp}), and $reg(s^h, \pi) = 0$ by the definition of the regret decomposition in Proposition~\ref{prop:creg}. This allows us to further simplify the expression to the following

\begin{multline}
	reg(s_0, \pi) = - V(s_0, \pi^*) +
	\sum_{a \in A} \pi(s_0, a) \cdot \bigg[\bar{C}(s_0, a)  +
	\sum_{s' \in S}T(s_0, a, s') \cdot \bigg[\sum_{a \in A}\pi(s', a) \cdot \bigg[\bar{C}(s', a)  + \ldots  \\
	\sum_{s^{h-1} \in S}T(s^{h-2}, a, s^{h-1}) \cdot \bigg[ \sum_{a \in A} \pi(s^{h\textnormal{-}1}, a) \cdot \bar{C}(s^{h\textnormal{-}1}, a)   \bigg] \ldots \bigg] \bigg] \bigg].
\end{multline}

The nested sum is simply the expected cost of the policy, $V(s_0, \pi)$. Thus, the expression can be further simplified to give the final result
	$reg(s_0, \pi) = V(s_0, \pi) - V(s_0, \pi^*)$, which is the original definition for the regret of a policy. $\qedsymbol$

\noindent
\section{Proof of Proposition~\ref{prop:equivalence}}

\begin{proposition}
	If uncertainties are independent per Def.~\ref{def:independence} then Problem~\ref{prob:1} is equivalent to  Problem~\ref{prob:creg_n1}.
\end{proposition}

In Problem~\ref{prob:1}, the agent first chooses a policy. The adversary observes the policy of the agent and reacts by choosing the uncertainty sample to be applied to maximise the regret for the policy of the agent. In Problem~\ref{prob:creg_n1}, the adversary reacts to the policy of the agent  by choosing the mapping from state-action pairs to uncertainty samples which maximises the regret for the policy of the agent. To prove the proposition, we show that in the case of independent uncertainty sets these adversaries are equivalent.

Let $\xi_{ind}$ be an independent uncertainty set. Then each sample of model uncertainty is $\xi_q = (C_q \in \mathcal{C}, T_q \in \mathcal{T}) \in \xi_{ind}$. By Definition~\ref{def:independence}, $\mathcal{T} = \times_{(s, a) \in S \times A} \mathcal{T}^{s, a} $ where $\mathcal{T}^{s, a}$ is the set of possible distributions over $S$ after applying $a$ in $s$, and $\mathcal{C} = \times_{(s, a) \in S \times A} \mathcal{C}^{s, a} $ where $\mathcal{C}^{s, a}$ is the set of possible expected costs of applying $a$ in $s$. We write $\times$ to denote the Cartesian product of sets.

The adversary in Problem~\ref{prob:creg_n1}, $\sigma^1: S \times A \times \xi \rightarrow \{0, 1\}$, maps each state and action chosen by the agent to an MDP sample such that the regret for the policy of the agent is maximised. This means that at a state-action pair, $s,a$, the adversary may apply any sample, $\xi^{s,a}_q = (C_{q}^{s,a} \in \mathcal{C}^{s,a}, T_{q}^{s,a} \in \mathcal{T}^{s,a})$, where we write $\xi^{s,a}_q$ to denote the MDP sample applied at $s,a$. At another state-action pair $s',a'$, the adversary may also apply any sample $\xi^{s',a'}_q = (C_{q}^{s',a'} \in \mathcal{C}^{s',a'}, T_{q}^{s',a'} \in \mathcal{T}^{s',a'})$, and so on. Thus, over all state-action pairs, the adversary may choose any combination of different samples at each state-action pair. We can write the set of all possible combinations of transition and cost functions as $\times_{(s, a) \in S \times A} \mathcal{T}^{s, a}$ and $\times_{(s, a) \in S \times A} \mathcal{C}^{s, a} $ respectively. These sets are equal to $\mathcal{T}$ and $\mathcal{C}$ respectively by the definition of independence. Thus, over all state-action pairs, the combination of samples chosen by $\sigma^1$ is equivalent to $\xi_q = (C_q \in \times_{(s, a) \in S \times A} \mathcal{C}^{s, a}, T_q \in \times_{(s, a) \in S \times A} \mathcal{T}^{s, a}) = (C_q \in \mathcal{C}, T_q \in \mathcal{T}) \in \xi_{ind}$, such that the regret for the policy of the agent is maximised. The adversary in Problem~\ref{prob:1} also chooses any $\xi_q = (C_q \in \mathcal{C}, T_q \in \mathcal{T}) \in \xi_{ind}$ such that the regret for the agent is maximised. Thus, we observe for independent uncertainties, the two adversaries maximising the regret are equivalent. Therefore $\argmin_{\pi \in \Pi} \max_{\xi_q \in \xi} reg_q(s_0, \pi) =   \argmin_{\pi \in \Pi} \max_{\sigma^1} reg(s_0, \pi).$ \qedsymbol

\section{Proof of Proposition~\ref{prop:bound_n1}}

\begin{proposition}
	\label{prop:bound_n1}
	If the expected number of steps for $\pi$ to reach $s_g\in G$ is at most $H$ for any adversary:
	\begin{equation*}
		0 \leq \max_{\sigma^{1}} reg(s_0, \pi) - \max_{\xi_q \in \xi} reg_q(s_0, \pi) \vspace{-1mm}\\ 
		\leq (\delta_C + 2\delta_{V^*}  + 2\delta_T C_{max}H )H, \ \ \textnormal{where}
	\end{equation*}

	\begin{align*}
		|\bar{C}_i(s, a) - \bar{C}_j(s, a)| & \leq \delta_C  & \forall s\in S, a \in A, \xi_i\in \xi, \xi_j \in \xi \\
		{\textstyle\sum}_{s'}|T_i(s, a, s') - T_j(s, a, s')| & \leq 2\delta_T & \forall s\in S, a \in A, \xi_i\in \xi, \xi_j \in \xi \\
		|V_i(s, \pi^*) - V_j(s, \pi^*)| & \leq \delta_{V^*} & \forall s\in S, \xi_i\in \xi, \xi_j \in \xi \\
		\bar{C}_i(s, a) & \leq C_{max} & \forall s\in S, a \in A, \xi_i\in \xi
	\end{align*}
\end{proposition}

For the lower bound, we observe that the adversary $\sigma^1$ can apply different combinations of $\xi_q \in \xi$ at each step, and therefore is more powerful than an adversary that chooses only a single $\xi_q \in \xi$. Therefore we have

\begin{equation}
	\max_{\sigma^1} reg(s_0, \pi) \geq \max_{\xi_q \in \xi} reg_q(s_0, \pi) \implies  \max_{\sigma^{1}} reg(s_0, \pi) - \max_{\xi_q \in \xi} reg_q(\pi) \geq 0.
\end{equation}

To prove the upper bound, we begin by denoting the most adversarial uncertainty sample by ${\xi_\top = (C_\top, T_\top) = \argmax_{\xi_q \in \xi} reg_q(s_0, \pi)}$. 
The corresponding value of regret for the most adversarial sample is denoted $reg_\top(s, \pi)$.
We introduce the following expression for the error in the maximum regret value at a given state

\begin{equation}
	\label{eq:diff_error}
    f(s) = \max_{\sigma^1} reg(s, \pi) - reg_\top(s, \pi).
\end{equation}

Thus we wish to find an upper bound for $f(s_0)$. We begin by expanding the expression for $f(s)$ using the regret decomposition from Proposition~\ref{prop:creg}

\begin{multline}
    f(s) = \max_{\sigma^1} reg(s, \pi) - reg_\top(s, \pi) \\
    = \max_{\xi_q \in \xi} \bigg[\sum_{a \in A} \pi(s, a) \cdot \bigg[ \bar{C}_q(s, a) 
    + \sum_{s' \in S} T_q(s, a, s')\cdot V_q (s', \pi^*)
    - V_q(s, \pi^*) + 
    \\
    \sum_{s' \in S} T_q(s, a, s') \cdot \max_{\sigma^1} reg(s', \pi) \bigg] \bigg]
    - \bigg[ \sum_{a \in A} \pi(s, a) \cdot \bigg[ \bar{C}_\top(s, a) 
    + \sum_{s' \in S} T_\top (s, a, s')\cdot V_\top(s', \pi^*) 
    \\
    - V_\top(s, \pi^*) + \sum_{s' \in S} T_\top(s, a, s')\cdot reg_\top(s', \pi) \bigg] \bigg].
\end{multline}

We can write the two max operators separately because when $n = 1$, the adversary can choose any $\xi_q \in \xi$ at each stage.
We introduce the following notation for the difference between the true worst-case sample and another sample
$$\Delta T_q(s, a, s') = T_q(s, a, s') - T_\top(s, a, s'),$$
$$\Delta V_q(s, \pi^*) = V_q(s, \pi^*) - V_\top(s, \pi^*).$$

We also make a substitution for $\max_{\sigma^1} reg(s', \pi)$ using the definition of $f(s')$ from Equation~\ref{eq:diff_error}. This results in the following expression
\begin{multline}
     f(s) =
    \max_{\xi_q \in \xi}
    \bigg[\sum_{a \in A} \pi(s, a) \cdot \bigg[\bar{C}_q(s, a) 
    + \sum_{s' \in S} (T_\top(s, a, s')
    + \Delta T_q(s, a, s'))\cdot
    (V_\top (s', \pi^*) +  \Delta V_q(s', \pi^*))
    - \\
     V_\top(s, \pi^*) - \Delta V_q(s, \pi^*) 
    + \sum_{s' \in S} (T_\top(s, a, s') + \Delta T_q(s, a, s')) \cdot
    (reg_\top(s', \pi) + f(s')) \bigg] \bigg] 
    - \\
    \bigg[\sum_{a \in A} \pi(s, a) \cdot \bigg[ \bar{C}_\top(s, a) + \sum_{s' \in S} T_\top (s, a, s')\cdot V_\top(s', \pi^*) - V_\top(s, \pi^*) + 
    \sum_{s' \in S} T_\top(s, a, s')\cdot reg_\top(s', \pi) \bigg ]\bigg].
\end{multline}

Cancelling terms and expanding, we have
\begin{multline}
     f(s) =
   \max_{\xi_q \in \xi}\bigg[ \sum_{a \in A} \pi(s, a) \cdot \bigg[\bar{C}_q(s, a) 
   +  \sum_{s' \in S}T_q(s, a, s')\cdot \Delta V_q(s', \pi^*) +
   \sum_{s' \in S}\Delta T_q(s, a, s')\cdot V_\top (s', \pi^*) 
   - \Delta V_q(s, \pi^*) +
   \\
    \sum_{s' \in S} T_q(s, a, s')\cdot f(s') + 
    \sum_{s' \in S}\Delta T_q(s, a, s')\cdot reg_\top(s', \pi) \bigg] \bigg] - \sum_{a \in A} \pi(s, a)\cdot \bar{C}_\top(s, a).
\end{multline}

At this point, we start to upper bound the terms in the previous equation using the constants defined in Proposition~\ref{prop:bound_n1}:

\begin{multline}
\label{eq:ref423}
f(s) \leq \delta_C + 2\delta_{V^*} + \max_{\xi_q \in \xi}\bigg[\sum_{a \in A} \pi(s, a) \bigg[ \sum_{s' \in S}\Delta T_q(s, a, s')\cdot V_\top (s', \pi^*)  + \\ \sum_{s' \in S} T_q(s, a, s')\cdot f(s') +
    \sum_{s' \in S}\Delta T_q(s, a, s')\cdot reg_\top(s', \pi) \bigg] \bigg].
\end{multline}

We observe that 

\begin{equation}
    \sum_{s' \in S}\Delta T_q(s, a, s')\cdot V_\top (s', \pi^*) \leq \delta_T \max_{s' \in S} V_\top (s', \pi^*) - \delta_T \min_{s' \in S} V_\top (s', \pi^*).
\end{equation}

Under the assumption that for policy $\pi$ the expected number of steps to reach the goal is at most $H$ for any adversary, the it holds that

\begin{equation}
   0 \leq V_\top (s', \pi^*) \leq V_\top (s', \pi) \leq C_{max}H.
\end{equation}

Therefore

\begin{equation}
    \sum_{s' \in S}\Delta T_q(s, a, s')\cdot V_\top (s', \pi^*) \leq \delta_T C_{max}H.
\end{equation}

By definition, $reg_\top(s, \pi) = V_\top(s, \pi) - V_\top(s, \pi^*)$, and can therefore similarly be bounded 

\begin{equation}
    0 \leq reg_\top(s', \pi) \leq C_{max}H,
\end{equation}

\begin{equation}
    \sum_{s' \in S}\Delta T_q(s, a, s')\cdot reg_\top (s', \pi^*) \leq \delta_T C_{max}H.
\end{equation}

Combining with Equation~\ref{eq:ref423} gives
\begin{equation}
\label{eq:state_bound}
f(s) \leq \delta_C + 2\delta_{V^*}  + 2\delta_T C_{max}H +  \max_{\xi_q \in \xi}\bigg[\sum_{a \in A} \pi(s, a) \cdot \bigg[ \sum_{s' \in S}T_q(s, a, s')\cdot f(s') \bigg] \bigg].
\end{equation}

Thus, we can write for the initial state

\begin{multline}
    f(s_0) \leq \delta_C + 2\delta_{V^*}  + 2\delta_T C_{max}H + \\
    \max_{\xi_q \in \xi}\bigg[\sum_{a \in A} \pi(s_0, a) \cdot \bigg[ \sum_{s' \in S\textbackslash G}T_q(s_0, a, s')\cdot f(s') +  \sum_{s_g' \in G}T_q(s_0, a, s_g')\cdot f(s_g') \bigg] \bigg ] \\
    \leq
     {\textstyle \Pr^{\pi, \sigma^1}}(\tau_{s_0}^G = 1)\cdot(\delta_C + 2\delta_{V^*}  + 2\delta_T C_{max}H) + \\
     {\textstyle \Pr^{\pi, \sigma^1}}(\tau_{s_0}^G \geq 2)\cdot \bigg[\delta_C + 2\delta_{V^*}  + 2\delta_T C_{max}H + \max_{s'\in S}f(s') \bigg],
\end{multline}

\noindent where in the last inequality we introduce the notation  $\Pr^{\pi, \sigma^1}(\tau_{s_0}^G = h)$ to denote the probability that, under $\pi$ and $\sigma^1$, a path starting from $s_0$ reaches a goal state in exactly $h$ steps, and observe that $f(s_g') = 0$ for all $s_g' \in G$. 

Using Equation~\ref{eq:state_bound} to substitute for $\max_{s' \in S}f(s')$ and repeatedly applying the same reasoning we have

\begin{equation}
    f(s_0) \leq (\delta_C + 2\delta_{V^*}  + 2\delta_T C_{max}H )\sum_{h=1}^\infty {\textstyle \Pr^{\pi, \sigma^1}}(\tau_{s_0}^G = h)\cdot h
\end{equation}

The sum on the right hand side is simply the expected number of steps to reach the goal and therefore we have the result

\begin{equation}
f(s_0) \leq (\delta_C + 2\delta_{V^*}  + 2\delta_T C_{max}H )H. \hspace{10pt}\qedsymbol
\end{equation}

\section{Proof of Proposition~\ref{prop:robust_equiv}}
\begin{proposition}
	Problem~\ref{prob:creg_n} is equivalent to finding the robust policy (Eq.~\ref{eq:robust}) for the $n$-UMDP.
\end{proposition}

To prove the proposition we show that the regret Bellman equation for an MDP in Proposition~\ref{prop:creg} is equivalent to the Bellman equation for the $n$-MDP. Therefore optimising minimax regret in the original UMDP according to Problem~\ref{prob:creg_n} is equivalent to optimising minimax expected cost on the $n$-UMDP (ie. finding the robust policy for the $n$-UMDP).

We start by considering the standard robust MDP problem introduced in Equation~\ref{eq:robust}.

\begin{equation*}
	\pi_{robust} = \argmin_{\pi \in \Pi} \max_{\sigma^1} V(s_0, \pi), \textnormal{where}
\end{equation*}
\begin{equation*}
	V(s, \pi) = \sum_{a \in A} \pi(s, a) \cdot  [\bar{C}(s,a) + \sum_{s' \in S}T(s, a, s') \cdot V(s', \pi)] .
\end{equation*}

Now consider solving the robust MDP problem on the $n$-UMDP. We need to apply the cost and transition functions from the $n$-UMDP, and replace actions by options. The adversary now changes the parameters after each option rather than each action. 

\begin{equation}
	\label{eq:robust_nmdp1}
	\pi_{robust}(n\textnormal{-}MDP) = \argmin_{\pi \in \Pi} \max_{\sigma^n} V(s_0, \pi), \textnormal{where}
\end{equation}
\begin{equation}
	\label{eq:robust_nmdp2}
	V(s, \pi^n) = 
	\sum_{o \in O} \pi^n(s, o) \cdot \bigg[C^o(s, o)
	+  \sum_{s' \in S}T(s, o, s') \cdot V(s', \pi^n) \bigg].
\end{equation}

To show that this is equivalent to solving Problem~\ref{prob:creg_n}, we start with the dynamic programming equation for the regret for a policy from Proposition~\ref{prop:creg}

\begin{equation*}
	reg(s, \pi) = \sum_{a \in A} \pi(s, a) \cdot \\ \Bigg[ \bar{C}(s, a) + \sum_{s' \in S}T(s, a, s') \cdot V(s', \pi^*)  - V(s, \pi^*) + \sum_{s' \in S}T(s, a, s')\cdot reg(s', \pi) \Bigg].
\end{equation*}

We unroll the definition for $n$ steps, and cancel terms in the same manner as in the proof of Proposition~\ref{prop:creg}. 

\begin{multline}
	reg(s, \pi) = 
	\sum_{a \in A} \pi(s, a) \cdot \bigg[\bar{C}(s, a) - V(s, \pi^*)  +
	\sum_{s' \in S}T(s, a, s') \cdot \bigg[\sum_{a \in A}\pi(s', a) \cdot \bigg[\bar{C}(s', a)  + \ldots  \\
	\sum_{s^{n\textnormal{-}1} \in S}T(s^{n\textnormal{-}2}, a, s^{n\textnormal{-}1}) \cdot \bigg[ \sum_{a \in A} \pi(s^{n\textnormal{-}1}, a) \cdot \bigg[ \bar{C}(s^{n\textnormal{-}1}, a) + 
	\sum_{s^{n} \in S}T(s^{n\textnormal{-}1}, a, s^{n}) \cdot V(s^n, \pi^*)
	+  \sum_{s^n \in S}T(s^{n\textnormal{-}1}, a, s^n) \cdot reg(s^n, \pi) \bigg] \bigg] \ldots \bigg] \bigg] \bigg].
\end{multline}

We can substitute policy $\pi$ for an equivalent option policy, $\pi^n$ which deterministically chooses an option in each state. At $s$, $\pi^n$ chooses a single option $o$. The associated policy $\pi^o$, executes the same action distribution as $\pi$ over the next $n$ steps.

\begin{multline}
	reg(s, \pi^n) = 
	\sum_{o \in O} \pi^n(s, o) \cdot \sum_{a \in A} \pi^o(s, a) \cdot \bigg[\bar{C}(s, a)  - V(s, \pi^*) +
	\sum_{s' \in S}T(s, a, s') \cdot \bigg[\sum_{a \in A}\pi^o(s', a) \cdot \bigg[\bar{C}(s', a)  + \ldots  \\
	\sum_{s^{n\textnormal{-}1}\in S}T(s^{n\textnormal{-}2}, a, s^{n\textnormal{-}1}) \cdot \bigg[ \sum_{a \in A} \pi^o(s^{n\textnormal{-}1}, a) \cdot \bigg[ \bar{C}(s^{n\textnormal{-}1}, a) + 
	\sum_{s^{n} \in S}T(s^{n\textnormal{-}1}, a, s^{n}) \cdot V(s^n, \pi^*)
	+  \sum_{s^n \in S}T(s^{n\textnormal{-}1}, a, s^n) \cdot reg(s^n, \pi^n) \bigg] \bigg] \ldots \bigg] \bigg] \bigg].
\end{multline}

By Definition~\ref{def:nmdp} of the $n$-MDP, the expected value of applying $\pi^o$ for $n$ steps is $V^n(s, \pi^o)$. We can also replace the nested transition functions with $T(s, o, s')$ which by Definition~\ref{def:nmdp} is the transition function for applying an option. Making these substitutions gives

\begin{multline}
	reg(s, \pi^n) = 
	\sum_{o \in O} \pi^n(s, o) \cdot \bigg[V^n(s, \pi^o) + \sum_{s' \in S}T(s, o, s') \cdot V(s', \pi^*) - V(s, \pi^*)
	+  \sum_{s' \in S}T(s, o, s') \cdot reg(s', \pi^n) \bigg].
\end{multline}

Substituting the cost function, $C^o(s, o)$ given by Equation~\ref{eq:cost_fn} we can now write Problem~\ref{prob:creg_n} as

\begin{equation*}
	\pi_{reg}^n = \argmin_{\pi \in \Pi} \max_{\sigma^n} reg(s_0, \pi^n), \textnormal{where}
\end{equation*}
\begin{equation*}
	reg(s, \pi^n) = 
	\sum_{o \in O} \pi^n(s, o) \cdot \bigg[C^o(s, o)
	+  \sum_{s' \in S}T(s, o, s') \cdot reg(s', \pi^n) \bigg].
\end{equation*}

We observe that this statement of Problem~\ref{prob:creg_n} is identical to the formulation of the robust policy for the $n$-UMDP in Equations~\ref{eq:robust_nmdp1}-\ref{eq:robust_nmdp2}. Therefore solving Problem~\ref{prob:creg_n} is equivalent to finding the robust policy on the $n$-UMDP.  \qedsymbol

\section{Proof of Proposition~\ref{prop:bound_genn}}
\begin{proposition}
	For dependent uncertainty sets, 
	\begin{equation*}
		\max_{\sigma^{n}} reg(s_0, \pi) - \max_{\xi_q \in \xi} reg_q(s_0, \pi) \geq 0 \hspace{10pt} \forall\ n \in \mathbb{N},
	\end{equation*}
	\begin{equation*}
		\min_{\pi^n} \max_{\sigma^{n}} reg(s_0, \pi^n) \geq \min_{\pi^{kn}} \max_{\sigma^{kn}} reg(s_0, \pi^{kn})\hspace{7pt} \forall\ n, k \in \mathbb{N}.
	\end{equation*}
\end{proposition}

For the first part (Equation~\ref{eq:upper_rpop}), we observe that the adversary $\sigma^n$ can apply different combinations of $\xi_q \in \xi$ at each $n$ step interval, and therefore is more powerful than an adversary that chooses only a single $\xi_q \in \xi$. Therefore we have

\begin{equation}
	\max_{\sigma^n} reg(s_0, \pi) \geq \max_{\xi_q \in \xi} reg_q(s_0, \pi) \implies  \max_{\sigma^{n}} reg(s_0, \pi) - \max_{\xi_q \in \xi} reg_q(\pi) \geq 0.
\end{equation}

 For the second part (Equation~\ref{eq:factor_increase}), we start with the expression for the regret decomposition for the $n$-step option MDP from Equation~\ref{eq:creg_as_cost}.

\begin{equation*}
reg(s, \pi^n) = \sum_{o \in O} \pi^n(s, o) \cdot [C^o(s,o) + \sum_{s' \in S}T^{o}(s, o, s') \cdot reg(s', \pi^n) ].
\end{equation*}

Taking the minimax, and then unrolling the regret expression over $k$ sequences of $n$ steps we have

\begin{multline}
\min_{\pi^n} \max_{\sigma^{n}} reg(s, \pi^n) = \min_{\pi^n} \max_{\xi_1 \in \xi} \bigg[\sum_{o \in O} \pi^n(s, o) \cdot \bigg[C^o_1(s,o) + \sum_{s^n \in S}T^{o}_1(s, o, s^n) \cdot \bigg[\ldots\\
+  \sum_{s^{(k-1)n} \in S}T^{o}_{k-1}(s^{(k-2)n}, o, s^{(k-1)n})\cdot \bigg[\max_{\xi_k \in \xi} \bigg[\sum_{o \in O}\pi^n(s^{(k-1)n}, o) \cdot \bigg[C^o_k(s^{(k-1)n},o) + \\
\sum_{s^{kn} \in S}T^{o}_k(s^{(k-1)n}, o, s^{kn})\cdot \max_{\sigma^n}reg(s^{kn}, \pi^n) \bigg]\bigg]\bigg] \ldots \bigg]\bigg]\bigg],
\end{multline}

\noindent where there is a separate maximisation over samples at every $n$ steps as the adversary may change the sample. If instead we were to restrict the adversary to change the parameters at every $kn$ steps the expression would be

\begin{multline}
\min_{\pi^n} \max_{\sigma^{kn}} reg(s, \pi^n) =  \min_{\pi^n} \max_{\xi_q \in \xi} \bigg[\sum_{o \in O} \pi^n(s, o) \cdot \bigg[C^o_q(s,o) + \sum_{s^n \in S}T^{o}_q(s, o, s^n) \cdot \bigg[\ldots\\
+  \sum_{s^{(k-1)n} \in S}T^{o}_{q}(s^{(k-2)n}, o, s^{(k-1)n})\cdot \bigg[\sum_{o \in O}\pi^n(s^{(k-1)n}, o) \cdot \bigg[C^o_q(s^{(k-1)n},o) + \\ \sum_{s^{kn} \in S} 
T^{o}_q(s^{(k-1)n}, o, s^{kn})\cdot \max_{\sigma^{kn}}reg(s^{kn}, \pi^n) \bigg] \bigg] \dots \bigg] \bigg] \bigg].
\end{multline}

We observe that for $kn$ step dependence, the adversary performs a single maximisation to choose one sample over the entire sequence of $kn$ steps. This is in contrast to $n$ step dependence, where the adversary is more powerful as it performs a separate maximisation for each of the sequences of $n$ steps. Recursively applying this observation we have that

\begin{equation}
\label{eq:prop5result1}
\min_{\pi^n} \max_{\sigma^{n}} reg(s, \pi^n) \geq \min_{\pi^n} \max_{\sigma^{kn}} reg(s, \pi^n)\hspace{10pt} \forall\ n, k \in \mathbb{N}.
\end{equation}

Additionally, we have that 

\begin{equation}
	\label{eq:prop5result2}
	\min_{\pi^n} \max_{\sigma^{kn}} reg(s, \pi^n) \geq
	\min_{\pi^{kn}} \max_{\sigma^{kn}} reg(s, \pi^{kn}) \hspace{10pt} \forall\ n, k \in \mathbb{N},
\end{equation}

\noindent which holds because option policies may be history-dependent. Therefore, a larger number of steps for the option policy, $kn \geq n$, means that each option may consider more of the history, resulting in a more powerful policy. 

Combining the inequalities in Equations~\ref{eq:prop5result1} and~\ref{eq:prop5result2} we have the required result. \qedsymbol

\section{Constraint Linearisation}
Here we describe the process of linearising the nonlinear equality constraints in the optimisation problem in Table~\ref{tab:optim_prob}. We use standard techniques from mathematical programming (see~\cite{williams2013model} for more details). In the case of deterministic policies, the model is solved exactly. For stochastic policies, a linear-piecewise approximation of the original constraints is required.

\textbf{Deterministic Policies}
If we assume that $\pi^o$ is deterministic, then each of the $\pi^o(s, a)$ variables is binary. 
In this case, we can linearise the constraints in Equations~\ref{eq:creg_end} and~\ref{eq:value_end} using a ``big M'' method.
Introducing additional variables denoted $c_q'$, Equation~\ref{eq:creg_end} is replaced by the constraints in Equations~\ref{eq:first_constr}-\ref{eq:last_constr}

\begin{align}
& \label{eq:first_constr} c_q(s, t) = {\textstyle\sum}_a  c_q'(s,a, t) & \forall s \in S^q_{\bar{s}, t}, \xi_q, t \leq n-1 \\
& c_q'(s,a,t) \leq c_q(s, a, t) & \forall s \in S^q_{\bar{s}, t}, a, \xi_q, t \leq n-1 \\
&c_q'(s,a,t) \leq \pi^o(s, a) \cdot M & \forall s \in S^q_{\bar{s}, t}, a, \xi_q, t \leq n-1 \\
& \label{eq:last_constr}  c_q'(s,a,t) \geq c_q(s, a, t) - (1-\pi^o(s,a))\cdot M & \forall s \in S^q_{\bar{s}, t}, a, \xi_q, t \leq n-1
\end{align}

$M$ is an upper bound on $c_q(s, a, t)$, and is domain-dependent. In a similar manner, Equation~\ref{eq:value_end} is replaced by equivalent constraints on additional variables $V_q^{n'}(s, a, t)$, where $M$ is chosen to be an upper bound on $V_q^n(s, a, t)$.

\textbf{Stochastic Policies}
In the case of stochastic policies, the $\pi^o(s, a)$ variables are continuous. The nonlinear constraints in Equation~\ref{eq:creg_end} can be approximated by applying separable programming. To convert the model into the appropriate form, we start by introducing additional variables. 

$$
x_q(s, a, t) = \frac{c_q(s, a, t) + \pi^o(s, a)}{2}
$$

$$
y_q(s, a, t) = \frac{c_q(s, a, t) - \pi^o(s, a)}{2}
$$

Equation~\ref{eq:creg_end} can now be written as:
\begin{equation}
c_q(s, t) = \sum_a \left[x_q(s, a, t)^2 - y_q(s, a, t)^2\right] \hspace{25pt} \forall s \in S^q_{\bar{s}, t}, \xi_q, t \leq n-1
\end{equation}

We apply the common technique from separable programming of approximating the quadratic terms by a piecewise linear function. By backwards induction, we can compute upper and lower bounds on $c_q(s, a, t)$, and therefore on $x_q(s, a, t)$ and $y_q(s, a, t)$. The range for $x_q(s, a, t)$ (and $y_q(s, a, t)$) is divided into $m$ intervals using $m+1$ breakpoints, $\{b_0, b_1, \ldots, b_m\}$. We introduce a variable, $\lambda_q^i(s, a, t)$ for each breakpoint, $i$. We then approximate $x_q(s, a, t)^2$ (and $y_q(s, a, t)^2$) with the following constraints

\begin{align}
& x_q(s, a, t)  = \sum_i \lambda_q^i(s, a, t)\cdot b_i & \forall s \in S^q_{\bar{s}, t}, a, \xi_q, t \leq n-1 \\
&x_q(s, a, t)^2  = \sum_i \lambda_q^i(s, a, t)\cdot b_i^2 & \forall s \in S^q_{\bar{s}, t}, a, \xi_q, t \leq n-1 \\
&\sum_i \lambda_q^i(s, a, t)  = 1 & \forall s \in S^q_{\bar{s}, t}, a, \xi_q, t \leq n-1 \\
&SOS2(\{\lambda_q^i(s, a, t)| 0 \leq i \leq m\}) & \forall s \in S^q_{\bar{s}, t}, a, \xi_q, t \leq n-1
\end{align}

\noindent where $SOS2$ indicates an adjacency constraint whereby at most two variables may have non-zero values, and if two variables are non-zero they must be adjacent. An identical process to that outlined above is applied to approximate and linearise the constraints in Equation~\ref{eq:value_end}.

In our experiments, we use 3 breakpoints for the piecewise linear approximation.

\section{Medical Decision Making Domain Details}
We adapt the medical decision making domain introduced by~\citet{sharma2019robust}. The state, $(h, d)  \in S$ comprises of 2 factors: the health of the patient, $h \in \{0, \ldots, 19\}$, and the day $d \in \{0, \ldots, 6\}$. At each state one of three actions can be applied, each representing different treatments. In each MDP sample the transition probabilities for each treatment differ, corresponding to different responses by patients with different underlying conditions. The health of the patient on the final day determines the cost received.

For each action, the possible relative changes in health are $h' - h = \Delta h \in \{-3, -2, -1, 0, 1, 2, 3 \}$. For each health level, a $3 \times 7$ matrix representing the likelihood of these 7 possible outcomes conditioned on each of the three actions was created randomly as follows. First, the nominal transition matrix for the UMDP is created by sampling 3 rows of a $7 \times 7$ identity matrix. Then, to create each MDP sample we add noise to the nominal transition values. Specifically, the absolute value of Gaussian zero-mean noise with standard deviation 0.1 was added to each value in the matrix, and the rows were then normalised to equal 1. 

The cost function is only applied according to the health state on the final day. The cost function was defined by ${C(h | d=6) =  0.05(19-h) + 2 \cdot \mathbbm{1}_{h=0} }$, where $\mathbbm{1}$ is the indicator function (a large penalty is added for reaching $h=0$).

\section{Disaster Rescue Domain Details}
We adapt this domain from the UMDP literature~\cite{adulyasak2015solving, ahmed2013regret, ahmed2017sampling, bagnell2001solving} to SSP MDPs. 
An agent navigates an 8-connected grid which contains swamps and obstacles by choosing from 8 corresponding actions.
Nominally, for each action the agent transitions to the corresponding target square with $p=0.8$, and to the two adjacent squares with $p=0.1$ each. 
If the target, or adjacent squares are obstacles, the agent transitions to that square with probability 0.05. 
Any remaining probability mass is assigned to not moving. 
If the square is a swamp, the cost is sampled uniformly in ${[1, 2]}$. The cost for entering any other state is 0.5. 
The agent does not know the exact locations of swamps and obstacles, and instead knows regions where they may be located.
To construct a UMDP, each square has a 1/15 chance of being the centre of a swamp region ($c_0$ and $c_1$ in Fig.~\ref{fig:disaster_rescue}), or obstacle region ($o_0$ in Fig.~\ref{fig:disaster_rescue}), respectively. 
The regions include the squares adjacent to the region centres (shaded areas in Fig.~\ref{fig:disaster_rescue}). 
To construct a sample, a swamp and obstacle is sampled uniformly from each swamp and obstacle region respectively.
Fig.~\ref{fig:disaster_rescue} (left) illustrates swamp and obstacle regions for a particular UMDP. 
Fig.~\ref{fig:disaster_rescue} (right) illustrates a possible sample corresponding to the same UMDP.

\section{Underwater Glider Navigation Domain Details}
Here we outline the process of creating a UMDP abstraction of underwater glider navigation using ocean current forecasts, based on the approach from~\cite{liu2018solution}. For each UMDP, we randomly sample a region of the Norwegian sea within the boundaries of $61.1^\circ$ to $61.2^\circ$ latitude and $4.5^\circ$ to $4.65^\circ$ longitude. The region is discretised into grid cells with a side length of $L$. Each grid cell is associated with a state in the MDP abstraction, $s$. We write $\mathbf{x}_s$ to denote the position of the centre of the grid cell associated with $s$. We can map any position to discrete space: $\mathbf{x} \rightarrow s$ if $||\mathbf{x} - \mathbf{x}_s||_\infty < L/2$. Each action corresponds to a heading direction for the glider. 

For each MDP sample, we repeat the following process to generate an MDP corresponding the time of the day in hourly intervals between 6am and 6pm. We denote the velocity of the glider relative to the water when taking action $a$ by $\mathbf{v}_g(a)$. The velocity of the ocean current at state $s$ is denoted $\mathbf{v}_c(s)$. This value is found by referring to the ocean current forecast for the appropriate time of day, which is available online.\footnote{\url{https://marine.copernicus.eu/}} The expected position of the glider after applying action $a$ in state $s$ is
$$
\mathbb{E} [\mathbf{x'} | s, a] = \mathbf{x}_s + (\mathbf{v}_g(a) + \mathbf{v}_c(s))\cdot \Delta t,
$$

\noindent where $\Delta t$ is the time between each time step in the MDP abstraction. The resulting position of the glider is also subject to noise, $\mathbf{d}(s, a)$. This noise is due to multiple sources such as: heading tracking error of the glider, environmental disturbances, and the glider not starting the action exactly at $\mathbf{x}_s$. We model $\mathbf{d}(s, a)$ with Gaussian noise with covariance matrix $\mathbf{\Sigma}$ for all states and actions. Thus,

$$\mathbf{x'} |_{s, a} \sim \mathcal{N}(\mathbb{E} [\mathbf{x'} | s, a], \mathbf{\Sigma})$$

The transition probabilities in the MDP abstraction can be found by integrating this distribution over each of the grid cells. In our experiments we use forecast data for May 1st 2020 and the following values for the abstraction:

\begin{itemize}
	\item $L$ = 500m
	\item $||\mathbf{v}_g(a)||_2 =$ 0.6m/s for all actions
	\item $\Delta t = 800s$
	\item $\mathbf{\Sigma} = \textnormal{diag}(150^2, 150^2)$
	\item There are 12 heading directions (actions) evenly spaced over $360^\circ$
\end{itemize}

\newpage
\section{p-Values for Experimental Results}
To assess the statistical significance of the difference in performance between each of the methods, we include a table of $p$-values for each of the methods. The top row of each table shows the mean and standard deviation of the normalised maximum regret for each of the methods across all of the runs. The remainder of the table includes $p$-values, which can be interpreted as follows. The $p$-value in the row of Method 1, and the column of Method 2 is the $p$-value for the hypothesis that Method 1 has better average performance than Method 2, calculated using a standard two-sample t-test. If the results for Method 1 do not have a better (lower) average than Method 2, then no $p$-value is included.

To compute $p$-values for the disaster rescue and glider domain where we tested a number of problem sizes, we first average the normalised maximum regret across all runs of all problem sizes, and compute the associated standard deviation. This mean and standard deviation is used to compute the $p$-values. We only include methods which solved all problem sizes within the 600s time limit. Similarly, for the medical domain we include methods which were within the 600s time limit.

\subsection{Disaster rescue domain}

\includegraphics[width=1.\linewidth]{Figures/legend_edit3.png}

\begin{table}[h]
	\resizebox{\textwidth}{!}{%
	\footnotesize
	\centering
	\setlength\tabcolsep{2pt}
	\renewcommand{\arraystretch}{1.2} 
	\begin{tabular}{|c|c|c|c|c|c|c|c|c|c|c|c|}
		\hline
		\footnotesize{Method}                 &  \includegraphics[scale=0.22]{Figures/regd1.png} & \includegraphics[scale=0.22]{Figures/regd2.png} & \includegraphics[scale=0.22]{Figures/regd3.png} & \includegraphics[scale=0.22]{Figures/cemrd1.png} & \includegraphics[scale=0.22]{Figures/cemrd2.png} & \includegraphics[scale=0.22]{Figures/cemrd3.png} & \includegraphics[scale=0.22]{Figures/bestpol.png}&
		\includegraphics[scale=0.22]{Figures/avgmdp.png}&
		\includegraphics[scale=0.22]{Figures/robust.png}&
		\includegraphics[scale=0.22]{Figures/regs1.png}&
		\includegraphics[scale=0.15]{Figures/cemrs1.png}  \\ \hline
		max reg   & 0.576; 0.25 & 0.340; 0.19 & \textbf{0.297}; 0.17 &  0.840; 0.25 & 0.633; 0.25 & 0.521; 0.24 & 0.647; 0.25 & 0.661; 0.28  &  0.578; 0.27 & 0.353; 0.18 & 0.695; 0.23 \\ \hline
	\includegraphics[scale=0.22]{Figures/regd1.png} & - & - & - & < 0.0001 & 0.017 & - & 0.004 & 0.001 & 0.47 & - & < 0.0001 \\ \hline
	\includegraphics[scale=0.22]{Figures/regd2.png} & < 0.0001 & - & - & < 0.0001 & < 0.0001 & < 0.0001 & < 0.0001 & < 0.0001 & < 0.0001 & 0.26 & < 0.0001 \\ \hline
	\includegraphics[scale=0.22]{Figures/regd3.png} & < 0.0001 & 0.014 & - & < 0.0001 & < 0.0001 & < 0.0001 & < 0.0001 & < 0.0001 &  < 0.0001 & 0.0015 & < 0.0001 \\ \hline
	\includegraphics[scale=0.22]{Figures/cemrd1.png} & - & - & - & - & - & - & - & - & - & - & - \\ \hline
	\includegraphics[scale=0.22]{Figures/cemrd2.png} & - & - & - & < 0.0001 & - & - & 0.30 & 0.16 & - & - & 0.008 \\ \hline
	\includegraphics[scale=0.22]{Figures/cemrd3.png} & 0.018 & - & - & < 0.0001 & < 0.0001 & - & < 0.0001 & < 0.0001 & 0.019  & - & < 0.0001 \\ \hline
	\includegraphics[scale=0.22]{Figures/bestpol.png} & - & - & - & < 0.0001 & - & - & - & 0.31 & - & - & 0.031 \\ \hline
	\includegraphics[scale=0.22]{Figures/avgmdp.png} & - & - &  -& < 0.0001 & - & - & - & - & - & - & 0.11 \\ \hline
	\includegraphics[scale=0.22]{Figures/robust.png} & - & - & - & < 0.0001 & 0.049 & -  & 0.007 & 0.003 & - & - & < 0.0001 \\ \hline
	\includegraphics[scale=0.22]{Figures/regs1.png} & < 0.0001 & - & - & < 0.0001 & < 0.0001 & < 0.0001 & < 0.0001 & < 0.0001 & < 0.0001 & - & < 0.0001 \\ \hline
	\includegraphics[scale=0.16]{Figures/cemrs1.png} & - & - & - & < 0.0001 & - & - &-  &  -& - & - & - \\ \hline
	\end{tabular}
}	
	\caption{Top row contains mean normalised maximum regret for disaster rescue domain across all problem sizes, in the format: mean; standard deviation. The remainder of the table contains $p$-values for the comparisons between each method. Methods which did not find a solution for all problem sizes within the 600s time limit are not included.  \label{tab:p_values_disaster}}
\end{table}

\begin{table}[h]
	\resizebox{\textwidth}{!}{%
		\footnotesize
		\centering
		\setlength\tabcolsep{2pt}
		\renewcommand{\arraystretch}{1.2} 
		\begin{tabular}{|c|c|c|c|c|c|c|c|c|c|c|c|}
			\hline
			\footnotesize{Method}                 &  \includegraphics[scale=0.22]{Figures/regd1.png} & \includegraphics[scale=0.22]{Figures/regd2.png} & \includegraphics[scale=0.22]{Figures/regd3.png} & \includegraphics[scale=0.22]{Figures/cemrd1.png} & \includegraphics[scale=0.22]{Figures/cemrd2.png} & \includegraphics[scale=0.22]{Figures/cemrd3.png} & \includegraphics[scale=0.22]{Figures/bestpol.png}&
			\includegraphics[scale=0.22]{Figures/avgmdp.png}&
			\includegraphics[scale=0.22]{Figures/robust.png}&
			\includegraphics[scale=0.22]{Figures/regs1.png}&
			\includegraphics[scale=0.15]{Figures/cemrs1.png}  \\ \hline
			max reg   & 0.601; 0.28 & 0.410; 0.22 & \textbf{0.348}; 0.20 & 0.826; 0.25 & 0.636; 0.26 & 0.528; 0.24 & 0.680; 0.28 & 0.643; 0.28 & 0.626; 0.31 & 0.389; 0.21 & 0.669; 0.26 \\ \hline
			\includegraphics[scale=0.22]{Figures/regd1.png} & - & - & - & < 0.0001 & - & -  & 0.004 & 0.081 & 0.22 & - & 0.010 \\ \hline
			\includegraphics[scale=0.22]{Figures/regd2.png} & <0.0001 & - & -  & <0.0001 & <0.0001 & <0.0001 & <0.0001 & <0.0001 & <0.0001 & - & <0.0001 \\ \hline
			\includegraphics[scale=0.22]{Figures/regd3.png} & <0.0001 & 0.003 & - & <0.0001 & <0.0001  & <0.0001 & <0.0001 & <0.0001 & <0.0001 & 0.031 &  <0.0001 \\ \hline
			\includegraphics[scale=0.22]{Figures/cemrd1.png} & - & - & - & - & - & - & - & - &-  &  -&  -\\ \hline
			\includegraphics[scale=0.22]{Figures/cemrd2.png} & - & - & - & <0.0001 & - & - & 0.064  & 0.40 & - & - & 0.12 \\ \hline
			\includegraphics[scale=0.22]{Figures/cemrd3.png} & 0.005 & - & - & <0.0001 & <0.0001 & - & <0.0001 & <0.0001 & <0.0001 & - & <0.0001 \\ \hline
			\includegraphics[scale=0.22]{Figures/bestpol.png} & - & - & - & <0.0001 & - & - & - & - & - & - & - \\ \hline
			\includegraphics[scale=0.22]{Figures/avgmdp.png} & - & - & - & <0.0001 & - & - & 0.11 & - & - & - & 0.18 \\ \hline
			\includegraphics[scale=0.22]{Figures/robust.png} & - & - & - & <0.0001 & 0.37 & - & 0.044 & 0.30 & - &-  & 0.080 \\ \hline
			\includegraphics[scale=0.22]{Figures/regs1.png} & <0.0001 & 0.18 & - & - & <0.0001 & <0.0001 & <0.0001 & <0.0001 & <0.0001 & - & <0.0001 \\ \hline
			\includegraphics[scale=0.16]{Figures/cemrs1.png} & - & - & - & <0.0001 & - & - & 0.35  & - & - & - & - \\ \hline
		\end{tabular}
	}	
	\caption{Top row contains mean normalised maximum regret for disaster rescue domain on the test set across all problem sizes, in the format: mean; standard deviation. The remainder of the table contains $p$-values for the comparisons between each method. Methods which did not find a solution for all problem sizes within the 600s time limit are not included.  \label{tab:p_values_disaster}}
\end{table}

\newpage
\FloatBarrier
\subsection{Underwater glider domain}

\includegraphics[width=1.\linewidth]{Figures/legend_edit3.png}

\begin{table}[h]
	\resizebox{\textwidth}{!}{%
		\footnotesize
		\centering
		\setlength\tabcolsep{2pt}
		\renewcommand{\arraystretch}{1.2} 
		\begin{tabular}{|c|c|c|c|c|c|c|c|c|c|c|c|}
			\hline
			\footnotesize{Method}                 &  \includegraphics[scale=0.22]{Figures/regd1.png} & \includegraphics[scale=0.22]{Figures/regd2.png} & \includegraphics[scale=0.22]{Figures/regd3.png} & \includegraphics[scale=0.22]{Figures/cemrd1.png} & \includegraphics[scale=0.22]{Figures/cemrd2.png} & \includegraphics[scale=0.22]{Figures/cemrd3.png} & \includegraphics[scale=0.22]{Figures/bestpol.png}&
			\includegraphics[scale=0.22]{Figures/avgmdp.png}&
			\includegraphics[scale=0.22]{Figures/robust.png}&
			\includegraphics[scale=0.22]{Figures/regs1.png}&
			\includegraphics[scale=0.15]{Figures/cemrs1.png}  \\ \hline
			max reg   & 0.590; 0.25 & 0.537; 0.22 & \textbf{0.504}; 0.21 & 0.813; 0.21 & 0.767; 0.20 & 0.696; 0.19 & 0.566; 0.23 &  0.652; 0.25 & 0.681; 0.27 & 0.557; 0.21 & 0.802; 0.20 \\ \hline
			\includegraphics[scale=0.22]{Figures/regd1.png}  & - & - & - & <0.0001 & <0.0001 & <0.0001 & - & 0.016 & 0.001 & - & <0.0001 \\ \hline
			\includegraphics[scale=0.22]{Figures/regd2.png} &  0.026 & - & - & <0.0001 & <0.0001 & <0.0001 & 0.13 & <0.0001 & <0.0001 & 0.21 &  <0.0001 \\ \hline
			\includegraphics[scale=0.22]{Figures/regd3.png} & 0.0007  & 0.092 & - & <0.0001 & <0.0001 & <0.0001 & 0.0075 & <0.0001 & <0.0001 & 0.015 & <0.0001 \\ \hline
			\includegraphics[scale=0.22]{Figures/cemrd1.png} & - & - & - & - &-  & - & - & - & - & - & - \\ \hline
			\includegraphics[scale=0.22]{Figures/cemrd2.png} & - & - & - & 0.026 & - & - & - & - & - & - & 0.065 \\ \hline
			\includegraphics[scale=0.22]{Figures/cemrd3.png} & - & - & - & <0.0001 & - & - & - & - & - & - & <0.0001 \\ \hline
			\includegraphics[scale=0.22]{Figures/bestpol.png} & 0.19 & - & - & <0.0001 & <0.0001 & <0.0001 & - & 0.001 & <0.0001 & - & - \\ \hline
			\includegraphics[scale=0.22]{Figures/avgmdp.png} & - & - & - & <0.0001 & <0.0001 & 0.044 & - & - & 0.17 & - & - \\ \hline
			\includegraphics[scale=0.22]{Figures/robust.png} & - & - & - & <0.0001 & 0.001 & 0.29 & - &-  & - & - &  -\\ \hline
			\includegraphics[scale=0.22]{Figures/regs1.png} & 0.11 & - & - & - & - & <0.0001 & 0.36 & 0.0002 &  -& - & - \\ \hline
			\includegraphics[scale=0.16]{Figures/cemrs1.png} & - & - & - & 0.32 & - & - & - & - &  -& - & - \\ \hline
		\end{tabular}
	}	
	\caption{Top row contains mean normalised maximum regret for underwater glider domain across all problem sizes, in the format: mean; standard deviation. The remainder of the table contains $p$-values for the comparisons between each method. Methods which did not find a solution for all problem sizes within the 600s time limit are not included.  \label{tab:p_values_disaster}}
\end{table}

\begin{table}[h!]
	\resizebox{\textwidth}{!}{%
		\footnotesize
		\centering
		\setlength\tabcolsep{2pt}
		\renewcommand{\arraystretch}{1.2} 
		\begin{tabular}{|c|c|c|c|c|c|c|c|c|c|c|c|}
			\hline
			\footnotesize{Method}                 &  \includegraphics[scale=0.22]{Figures/regd1.png} & \includegraphics[scale=0.22]{Figures/regd2.png} & \includegraphics[scale=0.22]{Figures/regd3.png} & \includegraphics[scale=0.22]{Figures/cemrd1.png} & \includegraphics[scale=0.22]{Figures/cemrd2.png} & \includegraphics[scale=0.22]{Figures/cemrd3.png} & \includegraphics[scale=0.22]{Figures/bestpol.png}&
			\includegraphics[scale=0.22]{Figures/avgmdp.png}&
			\includegraphics[scale=0.22]{Figures/robust.png}&
			\includegraphics[scale=0.22]{Figures/regs1.png}&
			\includegraphics[scale=0.15]{Figures/cemrs1.png}  \\ \hline
			max reg   & 0.715; 0.24 & 0.679; 0.23 & \textbf{0.652}; 0.23 & 0.849; 0.21 & 0.812; 0.20 & 0.757; 0.19 & 0.688; 0.24 & 0.718; 0.23  &  0.758; 0.24 & 0.680; 0.21 & 0.827; 0.19 \\ \hline
			\includegraphics[scale=0.22]{Figures/regd1.png} & - & - & - & <0.0001 & 0.0001 & 0.047 & - & 0.46 & 0.061 & - & <0.0001 \\ \hline
			\includegraphics[scale=0.22]{Figures/regd2.png} & 0.093 & - & - & <0.0001 & <0.0001 & 0.0008 & 0.37  & 0.072 & 0.002 & 0.48 & <0.0001 \\ \hline
			\includegraphics[scale=0.22]{Figures/regd3.png} & 0.011 & 0.16 & - & <0.0001 & <0.0001 & <0.0001 & 0.093 & 0.007 & <0.0001 & 0.14 & <0.0001 \\ \hline
			\includegraphics[scale=0.22]{Figures/cemrd1.png} & - & - & - & - & - & - & - & - & - & - & - \\ \hline
			\includegraphics[scale=0.22]{Figures/cemrd2.png} & - & - & - & 0.060 & - & - & - &-  & - &-  & 0.25 \\ \hline
			\includegraphics[scale=0.22]{Figures/cemrd3.png} & - & - & - & 0.0001 & 0.008 & - & - & - & 0.48 & - & 0.0008 \\ \hline
			\includegraphics[scale=0.22]{Figures/bestpol.png} & 0.17 & - & - & <0.0001 & <0.0001 & 0.004 & - & 0.13 & 0.006 & - & - \\ \hline
			\includegraphics[scale=0.22]{Figures/avgmdp.png} & - & - & - & <0.0001 & 0.0001 & 0.055 & - & - & 0.071 & - & <0.0001 \\ \hline
			\includegraphics[scale=0.22]{Figures/robust.png} & - & - & - & 0.0003 & 0.018 & - & - & - & - & - & 0.003 \\ \hline
			\includegraphics[scale=0.22]{Figures/regs1.png} & 0.090 & - & - & <0.0001 & <0.0001 & 0.0005 & 0.38 & 0.068 & 0.002 & - & <0.0001 \\ \hline
			\includegraphics[scale=0.16]{Figures/cemrs1.png} & - & - & - & 0.17 & - & - & - & - & - & - & - \\ \hline
		\end{tabular}
	}	
	\caption{Top row contains mean normalised maximum regret for the underwater glider domain on the test set across all problem sizes, in the format: mean; standard deviation. The remainder of the table contains $p$-values for the comparisons between each method. Methods which did not find a solution for all problem sizes within the 600s time limit are not included.  \label{tab:p_values_disaster}}
\end{table}

\newpage
\FloatBarrier
\subsection{Medical decision making domain}

\includegraphics[width=1.\linewidth]{Figures/legend_edit3.png}

\begin{table}[h]
	\resizebox{\textwidth}{!}{%
		\footnotesize
		\centering
		\setlength\tabcolsep{2pt}
		\renewcommand{\arraystretch}{1.2} 
		\begin{tabular}{|c|c|c|c|c|c|c|c|c|c|c|c|}
			\hline
			\footnotesize{Method}                 &  \includegraphics[scale=0.22]{Figures/regd1.png} & \includegraphics[scale=0.22]{Figures/regd2.png} & \includegraphics[scale=0.22]{Figures/regd3.png} & \includegraphics[scale=0.22]{Figures/cemrd1.png} & \includegraphics[scale=0.22]{Figures/cemrd2.png} & \includegraphics[scale=0.22]{Figures/cemrd3.png} & \includegraphics[scale=0.22]{Figures/bestpol.png}&
			\includegraphics[scale=0.22]{Figures/avgmdp.png}&
			\includegraphics[scale=0.22]{Figures/robust.png}&
			\includegraphics[scale=0.22]{Figures/regs1.png}&
			\includegraphics[scale=0.15]{Figures/cemrs1.png}  \\ \hline
			max reg & 0.596; 0.22 & 0.538; 0.18& $\mathbf{0.497}$; 0.16&0.906; 0.11&0.885; 0.13&0.880; 0.12&0.557; 0.20&0.596; 0.22&0.641; 0.23&0.529; 0.15&0.846; 0.12  \\ \hline
			\includegraphics[scale=0.22]{Figures/regd1.png} & - & - & - & <0.0001 & <0.0001 & <0.0001 & - & - & 0.013 & - & <0.0001 \\ \hline
			\includegraphics[scale=0.22]{Figures/regd2.png} & 0.0007 & - & - & - & - & - & 0.13 & 0.0007 & <0.0001 & - & <0.0001 \\ \hline
			\includegraphics[scale=0.22]{Figures/regd3.png} & - & 0.004 & - &-  & - & - & 0.0001 & <0.0001 & <0.0001 & - & <0.0001 \\ \hline
			\includegraphics[scale=0.22]{Figures/cemrd1.png} & - & - & - & - & - & - & - & - & - & - & - \\ \hline
			\includegraphics[scale=0.22]{Figures/cemrd2.png} & - & - & - & 0.025 & - & - & - & - & - & - & - \\ \hline
			\includegraphics[scale=0.22]{Figures/cemrd3.png} & - & - & - & 0.006 & 0.33 & - & - & - & - & - & - \\ \hline
			\includegraphics[scale=0.22]{Figures/bestpol.png} & 0.019 & - & - & - & - & - & - & 0.019 & <0.0001 & - & <0.0001 \\ \hline
			\includegraphics[scale=0.22]{Figures/avgmdp.png} & - & - & - & <0.0001 & <0.0001 & <0.0001 & - & - & 0.013 & - & <0.0001 \\ \hline
			\includegraphics[scale=0.22]{Figures/robust.png} & - & - & - & <0.0001 & <0.0001 & <0.0001 & - & - & - & - & <0.0001 \\ \hline
			\includegraphics[scale=0.22]{Figures/regs1.png} & <0.0001 & 0.27 & - & <0.0001 & <0.0001 & <0.0001 & 0.039 & <0.0001 & <0.0001 & - & <0.0001 \\ \hline
			\includegraphics[scale=0.16]{Figures/cemrs1.png} & - & - & - & <0.0001 & 0.0003 & 0.0008 & - & - & - & - & - \\ \hline
		\end{tabular}
	}	
	\caption{Top row contains mean normalised maximum regret for medical decision making domain over all 250 runs, in the format: mean; standard deviation. The remainder of the table contains $p$-values for the comparisons between each method. Methods which did not find a solution for all problem sizes within the 600s time limit are not included.  \label{tab:p_values_disaster}}
\end{table}

\begin{table}[h]
	\resizebox{\textwidth}{!}{%
		\footnotesize
		\centering
		\setlength\tabcolsep{2pt}
		\renewcommand{\arraystretch}{1.2} 
		\begin{tabular}{|c|c|c|c|c|c|c|c|c|c|c|c|}
			\hline
			\footnotesize{Method}                 &  \includegraphics[scale=0.22]{Figures/regd1.png} & \includegraphics[scale=0.22]{Figures/regd2.png} & \includegraphics[scale=0.22]{Figures/regd3.png} & \includegraphics[scale=0.22]{Figures/cemrd1.png} & \includegraphics[scale=0.22]{Figures/cemrd2.png} & \includegraphics[scale=0.22]{Figures/cemrd3.png} & \includegraphics[scale=0.22]{Figures/bestpol.png}&
			\includegraphics[scale=0.22]{Figures/avgmdp.png}&
			\includegraphics[scale=0.22]{Figures/robust.png}&
			\includegraphics[scale=0.22]{Figures/regs1.png}&
			\includegraphics[scale=0.15]{Figures/cemrs1.png}  \\ \hline
			max reg   & 0.674; 0.19 &  0.636; 0.18 & 0.625; 0.18 & 0.870; 0.13 &0.855; 0.14 &0.852; 0.13&0.706; 0.20& 0.677; 0.20 &0.715; 0.19 & $\mathbf{0.574}$; 0.12&0.814;0.13 \\ \hline
			\includegraphics[scale=0.22]{Figures/regd1.png} & - & - & - & <0.0001 & <0.0001 & <0.0001 & 0.034 & 0.43 & 0.008 & - & <0.0001 \\ \hline
			\includegraphics[scale=0.22]{Figures/regd2.png} & 0.011 & - & - & <0.0001 & <0.0001 & <0.0001 & <0.0001 & 0.008 & <0.0001 & - & <0.0001 \\ \hline
			\includegraphics[scale=0.22]{Figures/regd3.png} & 0.0016 & 0.25 & - & <0.0001 & <0.0001 & <0.0001 & <0.0001 & 0.0012 & <0.0001 & - & - \\ \hline
			\includegraphics[scale=0.22]{Figures/cemrd1.png} & - & - & - & - & - & - & - & - & - & - & - \\ \hline
			\includegraphics[scale=0.22]{Figures/cemrd2.png} & - & - & - & 0.11 & - & - & - & - & - & - & - \\ \hline
			\includegraphics[scale=0.22]{Figures/cemrd3.png} & - & - & - & 0.061 & 0.40 & - & - & - & - & - & - \\ \hline
			\includegraphics[scale=0.22]{Figures/bestpol.png} & - & - & - & <0.0001 & <0.0001 & <0.0001 & - & - & 0.30 & - & <0.0001 \\ \hline
			\includegraphics[scale=0.22]{Figures/avgmdp.png} & - & - & - & <0.0001 & <0.0001 & <0.0001 & 0.053 & - & 0.015 & - & <0.0001 \\ \hline
			\includegraphics[scale=0.22]{Figures/robust.png} & - & - & - & <0.0001 & <0.0001 & <0.0001 & - & - & - & - & <0.0001 \\ \hline
			\includegraphics[scale=0.22]{Figures/regs1.png} &  & <0.0001 & <0.0001 & 0.0001 & <0.0001 &<0.0001  & <0.0001 & <0.0001 & <0.0001 & - & <0.0001 \\ \hline
			\includegraphics[scale=0.16]{Figures/cemrs1.png} & - & - & - & <0.0001 & 0.0004 & 0.0009 & - & - & - & - & - \\ \hline
		\end{tabular}
	}	
	\caption{Top row contains mean normalised maximum regret on the test set for medical decision making domain over all 250 runs, in the format: mean; standard deviation. The remainder of the table contains $p$-values for the comparisons between each method. Methods which did not find a solution for all problem sizes within the 600s time limit are not included.  \label{tab:p_values_disaster}}
\end{table}

\end{appendices}

\end{document}